\documentclass[10pt,journal,compsoc]{IEEEtran}

\ifCLASSOPTIONcompsoc
  \usepackage[nocompress]{cite}
\else
  \usepackage{cite}
\fi

\ifCLASSINFOpdf
\else
\fi

\usepackage{enumitem}
\usepackage{amsmath}
\usepackage{amssymb}
\usepackage{algorithmic}
\usepackage{array}
\usepackage{subfigure}
\usepackage{epsfig}
\usepackage{graphicx}
\usepackage{multirow}
\usepackage{color}
\usepackage{threeparttable}
\usepackage{algorithmic}
\usepackage{algorithm}

\newtheorem{theorem}{Theorem}

\usepackage{fixltx2e}
\begin{document}
%
\title{Fast Batch Nuclear-norm Maximization and Minimization for Robust Domain Adaptation}
%
%
%

\author{Shuhao Cui, Shuhui~Wang,~\IEEEmembership{Member,~IEEE,} Junbao Zhuo,~
	Liang Li,\\~Qingming~Huang,~\IEEEmembership{Fellow,~IEEE,}
	and~Qi~Tian,~\IEEEmembership{Fellow,~IEEE}
	\IEEEcompsocitemizethanks{
		\IEEEcompsocthanksitem Corresponding author: Shuhui Wang.
		
		\IEEEcompsocthanksitem  S. Cui, S. Wang, J. Zhuo and L. Li are with the Key Laboratory of Intelligent Information Processing, Institute of Computing Technology, Chinese Academy of Sciences, Beijing 100190, China. S. Cui is also with the School of Computer Science and Technology, University  of  Chinese  Academy  of  Sciences, Beijing 101408, China. \protect
		E-mail: \{cuishuhao18s, wangshuhui, liang.li\}@ict.ac.cn, junbao.zhuo@vipl.ict.ac.cn.
		\IEEEcompsocthanksitem  Q. Huang is with the School of Computer Science and Technology, University  of  Chinese  Academy  of  Sciences, Beijing 101408, China, with the Key Laboratory of Intelligent Information Processing, Institute of Computing Technology, Chinese Academy of Sciences, Beijing 100190, China and also with Peng Cheng Laboratory, Shenzhen 518066, China. \protect
		E-mail: qmhuang@ucas.ac.cn.
		\IEEEcompsocthanksitem  Q. Tian is with Cloud BU, Huawei Technologies, Shenzhen 518129, China. \protect
		E-mail: tian.qi1@huawei.com. \protect}
} 

%
%

\markboth{Journal of \LaTeX\ Class Files,~Vol.~14, No.~8, August~2015}%
{Shell \MakeLowercase{\textit{et al.}}: Bare Demo of IEEEtran.cls for Computer Society Journals}
\IEEEdisplaynontitleabstractindextext


%

\IEEEtitleabstractindextext{%
	\begin{abstract}
		Due to the domain discrepancy in visual domain adaptation, the performance of source model degrades when bumping into the high data density near decision boundary in target domain. A common solution is to minimize the Shannon Entropy to push the decision boundary away from the high density area. However, entropy minimization also leads to severe reduction of prediction diversity, and unfortunately brings harm to the domain adaptation. 
		In this paper, we investigate the prediction discriminability and diversity by studying the structure of the classification output matrix of a randomly selected data batch. We find by theoretical analysis that the prediction discriminability and diversity could be separately measured by the Frobenius-norm and rank of the batch output matrix. The nuclear-norm is an upperbound of the former, and a convex approximation of the latter.
		Accordingly, we propose Batch Nuclear-norm Maximization and Minimization, which performs nuclear-norm maximization on the target output matrix to enhance the target prediction ability, and nuclear-norm minimization on the source batch output matrix to increase applicability of the source domain knowledge. We further approximate the nuclear-norm by $L_{1,2}$-norm, and design multi-batch optimization for stable solution on large number of categories. The fast approximation method achieves $O(n^2)$ computational complexity and better convergence property.
		Experiments show that our method could boost the adaptation accuracy and robustness under three typical domain adaptation scenarios. The code is available at https://github.com/cuishuhao/BNM.
	\end{abstract}
	}
\maketitle 
\IEEEdisplaynontitleabstractindextext
\IEEEpeerreviewmaketitle
\IEEEraisesectionheading{\section{Introduction}\label{s1}}
Deep neural networks have achieved great success in a wide range of computer vision applications. Nevertheless, it is well known that visual deep models require vast amounts of labeled data, which relies on intensive human efforts that are both expensive and time-consuming. The labeling cost becomes even more prohibitive for the instance level or even pixel level labeling. Without sufficient amount of labeled training samples, as a common consequence, spurious predictions will be made, even with only a subtle departure from the training samples. Actually, in most applications, there exists large domain discrepancy between source and target data. To reduce such notorious domain discrepancy, researchers resort to Domain Adaptation (DA), which aims to enable knowledge transfer from labeled source domain to an unlabeled target domain.
Recent advances in domain adaptation are mainly achieved by moment-alignment-based distribution matching~\cite{long2015learning,sun2016deep,yan2017mind} and adversarial learning~\cite{ganin2016domain,hoffman2017cycada,saito2018maximum}. 

\begin{figure}[t]
	\begin{center}
		\includegraphics[width=0.95\linewidth]{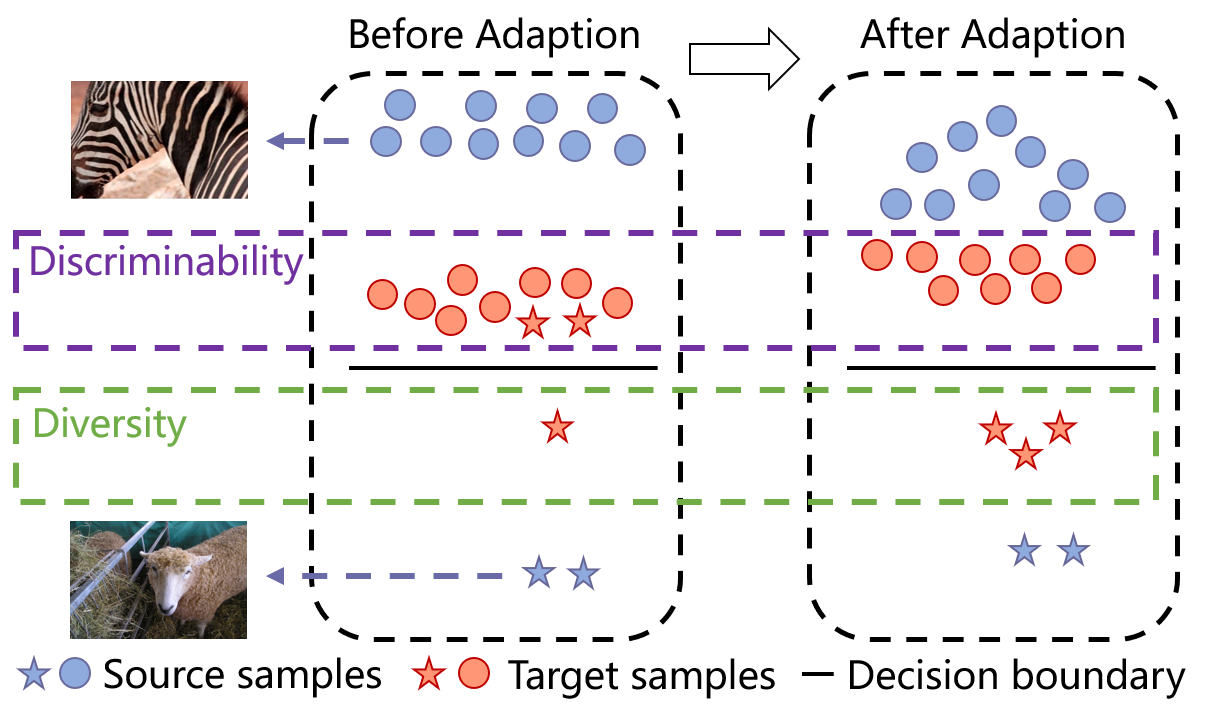}
	\end{center}
	\vspace{-2ex}
	\caption{Illustration of good transfer properties. Both prediction discriminability and diversity gain improvement after the adaptation procedure. More discriminability means more reliable predictions with less data near the decision boundary, as shown in the dotted purple frame. More diversity means more predictions on the minority categories, as shown in the dotted green frame.}
	\vspace{-1ex}
	\label{intro1}
\end{figure}

Among the endeavours in developing effective domain adaptation mechanism, the transfer properties such as prediction discriminability~\cite{residuallong,vu2019advent} and diversity \cite{zou2018unsupervised,zou2019confidence} are found to be the key to determine the overall cross-domain generalization ability and robustness, regardless of the backbone networks or transfer mechanisms used for model construction. Take the classification task as an example, the discriminability and diversity can be reflected by the expected model behaviour in the target domain that produces highly confident predictions on target domain examples from all the categories rather than only part of them.
To be more specific, in Figure~\ref{intro1}, as a widely accepted paradigm~\cite{joachims1999transductive}, higher prediction discriminability in target domain indicates predictions with more certainty. In other word, the data density near the decision boundary should be lower, so the samples are expected to be dragged away from the decision boundary after domain adaptation.
However, due to the constraints of domain alignment and uncertainty reduction~\cite{shannon1948mathematical,residuallong,vu2019advent}, the ambiguous target samples, including those from minority categories with smaller number of samples, tend to be pushed towards the nearby source samples that belong to the majority categories with larger number of labeled samples, {\it i.e.}, classified as the majority classes. 
This leads to severe degradation on the prediction diversity, and even worse, destruction of the output structure under extremely imbalanced target distribution.

However, these transfer properties were only partially explored in previous study. 
A straightforward way to strengthen prediction discriminability is to minimize the Shannon Entropy of the prediction outputs on target domain~\cite{residuallong,vu2019advent}. 
Nevertheless, directly applying Shannon entropy minimization also leads to the reduction of diversity~\cite{cui2020nnm}.
To encourage the diversity, balance constraints~\cite{SongSYLS18,zhuo2019unsupervised} is developed to equilibrate the output distribution among categories under the guidance of various types of distribution priors. When the distribution priors are unavailable, the predicted pseudo-labels are generated to approximate the statistical distribution~\cite{zou2018unsupervised,zou2019confidence} to encourage higher prediction probability on the minority categories, which seems less straightforward towards more accurate prediction model.




In this work, we investigate how to achieve robust domain adaptation by analyzing the discriminability and diversity on the predictions of source and target domains. We start by looking at the structure of classification output matrix of a randomly selected data batch from either source or target domain. We find through rigorous theoretical analysis that the discriminability and diversity can be measured by the Frobenius-norm and rank of the batch output matrix, respectively. The Frobenius-norm of a matrix is bounded by the its nuclear-norm, and the nuclear-norm is a convex approximation of the matrix rank.
Accordingly, maximizing/minimizing the nuclear-norm leads to large/small Frobenius-norm and rank of the batch matrix, which can substantially regulate the discriminability and the diversity of the prediction output.

Based on the above theoretical findings, we propose Batch Nuclear-norm Maximization and Minimization~(BNM$^2$), a unified learning framework for visual domain adaptation. By looking into the two ends of the domain adaptation process, our method works on the prediction output matrices of both source and target domains. 
First, the output matrix nuclear-norm of a target domain data batch can be maximized straightforwardly to enhance both the prediction discriminability and diversity after domain adaptation, {\it a.k.a}, the Batch Nuclear-norm Maximization (BNM)~\cite{cui2020nnm}. 
Furthermore, due to complex factors such as overfitting, distribution bias and noise, the domain-specific knowledge learned in source domain might be difficult to transfer to the target domain, especially when facing with large domain discrepancy. 
Accordingly, we formulate the mechanism of Batch Nuclear-norm Minimization on source domain, to soften the source domain prediction outputs and make it more transferable. The nuclear-norm maximization and minimization play complementarily to elegantly adjust the discriminability and diversity of the visual models, and they can be combined with a wide range of domain adaptation frameworks.  






We further address the computational issues of BNM$^2$. 
First, SVD is typically employed to calculate the nuclear-norm in BNM~\cite{cui2020nnm}, but it might not be converged in some cases, see Table~\ref{SSDA}. We theoretically prove that the nuclear-norm
can be approximated by the main components of $L_{1,2}$-norm (see Sec. 1.4 in Supp). The approximation enables fast convergence of matrix computation, and speeds up the training with reduced computation complexity from $O(n^3)$ to $O(n^2)$.
Second, due to the limited batch size used in training, the nuclear-norm optimization tends to perform less effective on data domain with large number of categories. To deal with this case, we records multiple previous prediction batch matrices, and estimate nuclear-norm on the concatenated batch predictions to achieve more stableness.
Extensive experiments validate the effectiveness of our fast BNM$^2$ (FBNM$^2$) under diverse DA scenarios.


This journal paper is a systematic extension of our preliminiary work \cite{cui2020nnm}. From methodology aspect, beyond the nuclear-norm maximization on target domain~\cite{cui2020nnm}, we formulate a novel Batch Nuclear-norm Minimization mechanism on source domain. The batch nuclear-norm maximization on target domain and minimization on source domain can collaborate elegantly and combine with various types of domain adaptation frameworks.
From the computation aspect, we use $L_{1,2}$-norm to approximate the nuclear-norm and develop Multi-batch Nuclear-norm Optimization, while in~\cite{cui2020nnm} only the naive SVD is employed. Thus better efficiency and more stable solution is gained over ~\cite{cui2020nnm}.
From the experiment aspect, we construct a balanced dataset selected from Domainnet under unsupervised DA scenario. We extend our methods to semi-supervised DA, and insert our methods to recent frameworks such as HDAN~\cite{cui2020hda} and SHOT~\cite{liang2020we}. Details can be found in summary of change.
Our contributions are summarized as follows:
\begin{itemize}
	\item We theoretically prove that the discriminability and diversity of the prediction output can be jointly approximated by nuclear-norm of the batch output matrix.
	\item Based on the theoretical findings, we propose a new DA paradigm BNM$^2$, which achieves better prediction discriminability and diversity.
	\item We develop fast BNM$^2$, which achieves $O(n^2)$ computational complexity and more stable solution. 
	\item Extensive experiments validate the effectiveness of the proposed solution and the flexibility to cooperate with existing methods.
\end{itemize}

\section{Related work}
\label{s2}

Visual domain adaptation~\cite{saenko2010adapting} has gained significant improvement in the past few years. Recently, the study of DA methods focuses on diverse adaptation circumstances, such as unsupervised DA~\cite{ganin2016domain,Xu_2019_ICCV,cui2020nnm}, multi-source DA~\cite{Xu_2018_CVPR,peng2019moment}, semi-supervised DA~\cite{saito2019semi} and unsupervised open DA~\cite{zhuo2019unsupervised}. 
Similar to existing methods of \cite{Xu_2019_ICCV,cui2020hda,versatile}, we explore the transfer properties in DA, to construct general frameworks suitable for various DA scenarios.

In terms of distribution alignment, deep domain adaptation methods incorporate two main technologies into deep networks, {\it i.e.}, moment alignment and adversarial training. Moment alignment~\cite{gretton2012kernel,long2015learning,yan2017mind} are devised with hand-crafted metrics, such as maximum mean discrepancy~\cite{long2015learning,long2017deep}, second-order statistics~\cite{sun2016deep,zhuo2017deep} or other distance metrics on the representations~\cite{kang2018deep}. Pioneered by Generative Adversarial Networks (GAN) \cite{goodfellow2014generative}, adversarial learning has been successfully explored on various tasks including domain adaptation. Domain adversarial neural network (DANN)~\cite{ganin2016domain}, confuses the domain classifier by the gradient reversal layer, to lessen the domain shift. CyCADA \cite{hoffman2017cycada} and CDAN \cite{long2018conditional} further devise adversarial frameworks inspired by powerful GANs. For individual domain adaptation, ADDA \cite{tzeng2017adversarial}, MCD \cite{saito2018maximum} and GVB~\cite{cui2020gvb} construct effective structures from the perspective of the minmax game.

Apart from distribution alignment, some methods optimize prediction discriminability and diversity by self training~\cite{Zou_2019_ICCV,qi2021self} on target domain. To increase discriminability, Shannon Entropy~\cite{shannon1948mathematical} is directly minimized in~\cite{residuallong,vu2019advent} to obtain more determined predictions for target samples. Entropy minimization is further modified into maximum square loss in \cite{chen2019domain}, to reduce the influence of easy-to-transfer samples. Methods in \cite{hu2018duplex,liang2020we} calculate pseudo-labels on target domain, and further improve prediction discriminability by minimizing cross-entropy loss with pseudo-labels. Meanwhile, prediction discriminability is supposed to be lessened in~\cite{BSP_ICML_19} towards more transferability by penalizing the main components of the features. Higher prediction transferability is also achieved by smooth labeling in~\cite{mullerdoes}. By harmonizing prediction transferability and discriminability, SAFN~\cite{Xu_2019_ICCV} is proposed by increasing the feature norm in target domain.


To maintain prediction diversity, a direct technique is imbalanced learning~\cite{he2009learning}. Typical methods such as \cite{ubias,zhuo2019unsupervised} directly enforce the ratio of predictions on minority categories. However, the category distribution is required as prior knowledge. Without the prior knowledge, the prediction results are adopted as pseudo-labels in \cite{zou2019confidence,zou2018unsupervised,Zou_2019_ICCV} to approximate the category distribution. Method in \cite{liang2020we} increases prediction diversity by maximizing the mutual information between intermediate feature representations and
classifier outputs. From another technical aspect, Determinantal Point Processes (DPPs)~\cite{kulesza2012determinantal} act probabilistically to capture the balance between quality and diversity within sets, but suffer from the large computation complexity. It worths emphasizing that diversity is a more fundamental problem than class imbalance. First, the prediction diversity might still degrade even under balanced situations, as discussed in Table~\ref{batchsamplesstatic}. Second, category imbalance is a problem on common datasets, while diversity is a desired property on the predictions and generations. Ensuring diversity is a solution for a wide range of machine learning models that avoids the notorious mode collapse phenomenon~\cite{arjovsky2017wasserstein}.

In this paper, we study the domain adaptation process from the perspective of matrix analysis~\cite{candes2009exact,cai2010singular}, which has already been widely applied to numerous computer vision tasks, such as image denoising~\cite{gu2014weighted} and image restoration~\cite{dong2012nonlocal}. Among these tasks, a common assumption is that the noisy data brings extra components to the matrix. To reduce the influence of the extra components, minimizing nuclear-norm of the matrix has been widely accepted in \cite{gu2014weighted,dong2012nonlocal}. In comparison to the above methods, we aim to explore and exploit the extra components in either source and target domain. The nuclear-norm of the target/source response matrix is maximized/minimized to regulate the domain adaptation process, and the nuclear-norm maximization on target response tends to play dominantly in achieving high discriminability and diversity. Recently, similar research is conducted on feature level, {\it e.g.}, BSP~\cite{BSP_ICML_19} penalizes the largest singular values in the feature matrix to boost prediction discriminability. In comparison, we directly work on the batch classification response matrix to adjust the discriminability and diversity.

\section{Method}
\begin{figure}
	\begin{center}
		\begin{minipage}{.98\textwidth}
			\begin{minipage}{.24\textwidth}
				\subfigure[Source only]{	
					\includegraphics[width=0.9\textwidth]{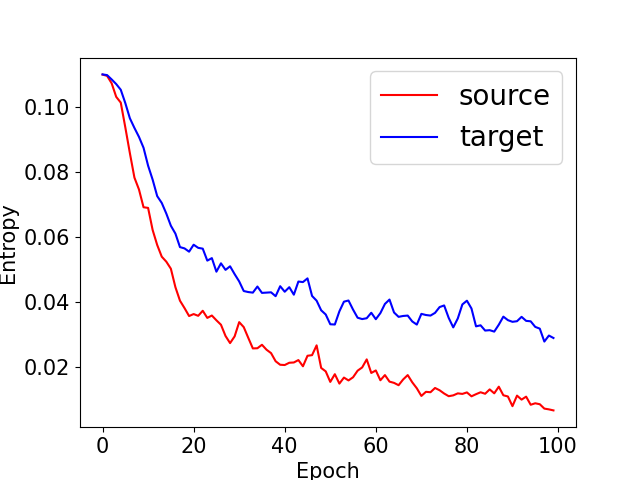}
					\label{afnnorm0}}
			\end{minipage}
			\begin{minipage}{.24\textwidth}
				\subfigure[SAFN]{
					\includegraphics[width=0.9\textwidth]{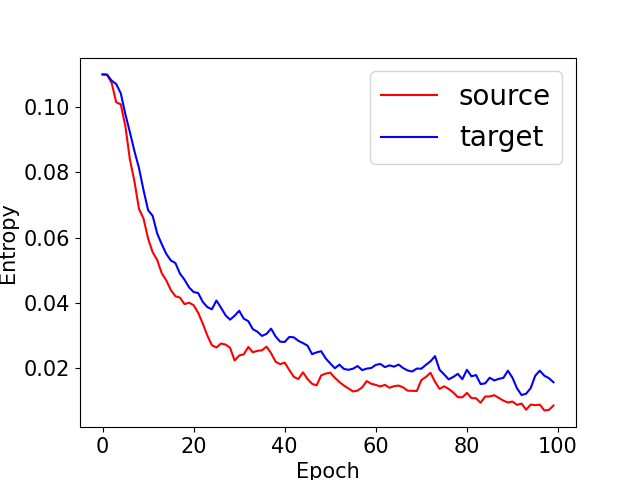}
					\label{afnnorm1}}
			\end{minipage}
			
			\vspace{-2ex}
			

		\end{minipage}
				\caption{The Entropy statistics during the whole training process for scenarios of Source only or SAFN. Low Entropy means high prediction discriminability for the corresponding method.}
	\label{afnnorm}
	\end{center}
	\vspace{-2ex}
\end{figure}

\subsection{On Discriminability and Diversity}

More prediction discriminability means less ambiguous target samples near the task-specific decision boundary. The ambiguous samples tend to be misclassified, thus increased prediction discriminability could ensure reliable predictions. Meanwhile, we reanalyze popular method of SAFN~\cite{Xu_2019_ICCV} as shown in Figure~\ref{afnnorm}. The blue line of SAFN indicates that the intrinsic effect of SAFN is to increase the feature norm towards more prediction discriminability. Detailed Discussion can be found in Sect. 2.1 in Supplementary, from both theoretical and practical perspectives. The results indicate that enhancing prediction discriminability is necessary in domain adaptation.


It is normal that a randomly selected batch with $B$ samples to be imbalanced, even they are sampled from a balanced dataset. 
To analyze this phenomenon, we construct a balanced dataset with categories $C=126$, {\it i.e.}, the Balanced Domainnet dataset, where each class has equal number of samples, as described in Sect. 4.2. On this dataset, we show the ratio of categories with different number of samples under different batch size $B$ in Table~\ref{batchsamplesstatic}, where each of the results is the average of three random batch sampling trials. For example, when $B$ equals $C$, more than one third of categories are not included in the batch. Predictions on these unsampled categories, denoted as minority categories, tend to be compromised during the model update step using this data batch. Similarly, under the same setting, only $7.8\%$ of the categories attain more than $3$ samples in the batch, denoted as majority categories. The majority categories tend to dominate the prediction output matrices. Even when $B=2C$, there are still $12.9\%$ categories unsampled in the batch, which validates the degradation on prediction diversity in balanced dataset would likely occur. 
In this case, directly increasing discriminability by Entropy minimization can only push the ambiguous samples to the majority categories. The continuous convergence to the majority categories results in inappropriate encouragement on wrong predictions and lessened prediction diversity. 


To investigate the discriminability and diversity in domain adaptation, we start by looking at the prediction outputs on a data batch with $B$ randomly selected samples. We denote the number of categories as $C$, and the batch prediction output matrix as $\mathrm{A} \in \mathbb{R}^{B \times C}$, which satisfies:
\begin{equation}
	\begin{split}{\mathrm
			\sum_{j=1}^C\mathrm{A}_{i,j}=1, \quad &\forall i \in 1...B \\
			\mathrm{A}_{i,j} \ge 0, \quad &\forall i \in 1...B, j \in 1...C.
	}\end{split}
	\label{matrix}
\end{equation}
It should be noted that popular deep methods can obtain well-structured response matrix $\mathrm{A}$, trained with large number of labeled source samples.

\begin{table}[t]
	\begin{center}
		\vspace{-5pt}
		\caption{Average ratio(\%) of categories with different number of samples in the batch.}
		\vspace{-2pt}
		\label{batchsamplesstatic}
		\setlength{\tabcolsep}{2mm}{
			\resizebox{0.35\textwidth}{!}{%
			\begin{tabular}{ccccc}
				\hline
				\#Samples & 0 & 1 &2 &$\geq$ 3  \\
				\hline\hline
				$B=0.5C$ &\textbf{60.4} &30.7 &7.5 &\textbf{1.4} \\
				$B=C$ & \textbf{36.4} &37.2 &18.6 &\textbf{7.8} \\
				$B=2C$ & \textbf{12.9} &27.2 &27.6 &\textbf{32.3}\\
				\hline
		\end{tabular}}}
		\vspace{-10pt}
	\end{center}
\end{table}
\subsubsection{Discriminability Measurement: $F$-norm}
\label{fnorm}

Higher discriminability means lower prediction uncertainty in the response matrix $\mathrm{A}$. A natural measurement of the uncertainty is Shannon Entropy~\cite{shannon1948mathematical}, denoted as entropy for simplicity, which could be calculated as follows:
\begin{equation}
	\begin{split}{
			{H}(\mathrm{A}) = -\frac{1}{B}{\sum_{i=1}^B \sum_{j=1}^C \mathrm{A}_{i,j}\log(\mathrm{A}_{i,j}) }.
	}\end{split}
	\label{trace}
\end{equation}
For more discriminability, we could directly minimize $H(\mathrm{A})$, the same as~\cite{grandvalet2005semi,residuallong,vu2019advent}. When $H(\mathrm{A})$ reaches the minimum, only one entry is $1$ and the other $C-1$ entries are $0$ in each row of $\mathrm{A}$, {\it i.e.}, $\mathrm{A}_{i,j} \in \{0,1\} \quad \forall i \in 1...B, j \in 1...C$. The minimum of $H(\mathrm{A})$ indicates the highest prediction discriminability of $\mathrm{A}$, where the prediction $\mathrm{A}_i$ on each sample in the batch is fully determined.

To enforce $\mathrm{A}$ to approximate the solution with minimal $H(\mathrm{A})$, other loss functions could also be utilized. In this paper, We choose Frobenius-norm ($F$-norm) $\left\| \mathrm{A} \right\|_F$:
\begin{equation}
	\begin{split}{
			\left\| \mathrm{A} \right\|_F = \sqrt{\sum_{i=1}^B \sum_{j=1}^C |\mathrm{A}_{i,j}|^2 }.
	}\end{split}
	\label{trace}
\end{equation}

\begin{theorem}
Given matrix $\mathrm{A}$, $H(\mathrm{A})$ and $\left\|\mathrm{A}\right\|_F$ have strict opposite monotonicity about the values of matrix $\mathrm{A}$.
\end{theorem}
Detailed proof can be found in Sect. 1.1 in Supplementary. According to the theorem, the minimum of $H(\mathrm{A})$ and the maximum of $\left\| \mathrm{A} \right\|_F$ could be achieved at the same solution of $\mathrm{A}$. Particularly, the upper-bound of $\left\| \mathrm{A} \right\|_F$ can be calculated according to inequality of arithmetic and geometric means:
\begin{equation}
	\begin{split}{
			\left\| \mathrm{A} \right\|_F &\leq \sqrt{\sum_{i=1}^B (\sum_{j=1}^C \mathrm{A}_{i,j}) \cdot (\sum_{j=1}^C \mathrm{A}_{i,j})}\\
			&= \sqrt{\sum_{i=1}^B 1 \cdot 1 }\quad= \sqrt{B}.
	}\end{split}
	\label{tracemax}
\end{equation}
When the upper-bound of $\left\| \mathrm{A} \right\|_F$ and the minimum of $H(\mathrm{A})$ are simultaneously reached, the highest prediction discriminability would be attained at this data batch. 

Meanwhile, we could minimize $\left\| \mathrm{A} \right\|_F$ towards lessened prediction discriminability. The lower-bound of $\left\| \mathrm{A} \right\|_F$ could be calculated as:
\begin{equation}
	\begin{split}{
			\left\| \mathrm{A} \right\|_F &\geq \sqrt{\sum_{i=1}^B \frac{1}{C} \cdot(\sum_{j=1}^C \mathrm{A}_{i,j}) \cdot (\sum_{j=1}^C \mathrm{A}_{i,j})}\\
			&= \sqrt{\sum_{i=1}^B \frac{1}{C} \cdot 1 \cdot 1 }\quad= \sqrt{\frac{B}{C}},
	}\end{split}
	\label{tracemin}
\end{equation}
where the minimum could be reached when $\mathrm{A}_{i,j}$ equals $\frac{1}{C}$ for each $i$ and $j$. The circumstance satisfies the same prediction probabilities for all the $C$ categories, corresponding to the situation of lowest prediction discriminability .

\subsubsection{Diversity Measurement: Matrix Rank}
\label{rank}

Higher prediction diversity means more involved categories with non-zero values in matrix $\mathrm{A}$. For a certain dataset, the number of predicted categories in $\mathrm{A}$ is expected to be a constant on average, denoted as $\mathrm{E}_{C}$. To provide intuitive comprehension of $\mathrm{E}_{C}$, we construct a toy dataset, with only $2$ samples separately belonging to $2$ categories. We randomly select $2$ samples, and the samples hold $0.5$ probability of belonging to the same category, and $0.5$ probability of belonging to two different categories. Then the constant $\mathrm{E}_{C}$ could be calculated by:
\begin{equation}
	\begin{split}
	\mathrm{E}_{C} = 0.5 \times 1 + 0.5 \times 2 = 1.5,
    \end{split}
\end{equation}
where the matrix is expected to contain $1.5$ categories on average.


Generally speaking, the predicted constant $\mathrm{E}_{C}$ is the expectation of number of categories $C_p(\mathrm{A})$ for multiple $\mathrm{A}$ selected from domain $\mathcal{D}$, as follows: 
\begin{equation}
	\begin{split}
		\mathrm{E}_{C} &= \mathbb{E}_{\mathrm{A} \sim \mathcal{D}}(C_p(\mathrm{A} )).
	\end{split}
\end{equation}
For well-performed predictions, $\mathrm{E}_{C}$ should be similar to the average ground truth number of categories for all $\mathrm{A}$. 
Under certain application scenarios, the ground truth number of categories is always a constant. 
Thus the predicted number of categories $\mathrm{E}_{C}$ should also be a constant. If $\mathrm{E}_{C}$ becomes larger, more categories are involved, which corresponds to higher prediction diversity. 

Thus, for existing batch matrix $\mathrm{A}$, prediction diversity could be measured by $C_p(\mathrm{A})$, calculated by counting the number of predicted categories on the one-hot matrix, which is accomplished by the argmax operation on the prediction scores. 
The number of independent components of this matrix is the matrix rank:
\begin{equation}
	\begin{split}{
C_p(\mathrm{A})= rank(\mathbb{I}[ \mathrm{A}_{i,\arg \max(A_i)}]).
	}\end{split}
	\label{category}
\end{equation}

However, due to its discrete nature, the argmax operation could not facilitate the gradient propagation of the loss functions. Thus we further analyze the relationship between number of categories $C_p(\mathrm{A})$ and the predicted vectors in $\mathrm{A}$. Two randomly selected prediction vectors, {\it i.e.}, $\mathrm{A}_i$ and $\mathrm{A}_k$, could be linearly independent when $\mathrm{A}_i$ and $\mathrm{A}_k$ belong to different categories. When $i$ and $k$ are classified as the same category and $\left\| \mathrm{A} \right\|_F$ is near $\sqrt{B}$, the difference between $\mathrm{A}_i$ and $\mathrm{A}_k$ is tiny. Then $\mathrm{A}_i$ and $\mathrm{A}_k$ could be regarded as approximately linearly dependent. The largest number of linear independent vectors is called the matrix rank. Thus, $rank(\mathrm{A})$ could be an approximation of $C_p(\mathrm{A})$, if $\left\| \mathrm{A} \right\|_F$ is near the upper-bound $\sqrt{B}$, as follows:
\begin{equation}
	\begin{split}{
			C_p(\mathrm{A}) \approx  rank(\mathrm{A})
	}\end{split}
	\label{category}
\end{equation}

Apparently, the maximum value of $C_p(\mathrm{A})$ is $\min(B, C)$. When $B \geq C$, the maximum value is $C$, which firmly guarantees that the prediction diversity for the batch achieves the maximum. However, when $B < C$, the maximum value is less than $C$, maximization of $C_p(\mathrm{A})$ still enforces that the predictions on the batch samples should be as diverse as possible, though there is no guarantee that all the categories will be assigned to at least one example. Therefore, maximization of $C_p(\mathrm{A})$ could ensure the diversity in any case.

\subsection{Batch Nuclear-norm Calculation}
\label{theory}

\subsubsection{Discriminability and Diversity in Batch Nuclear-norm}

For a normal matrix, the calculation of the matrix rank is an NP-hard non-convex problem. Thus we could not directly restrain the rank of matrix $\mathrm{A}$. To explore the diversity in matrix $ \mathrm{A} $, we are supposed to find the relationship between $rank(\mathrm{A})$ and nuclear-norm.
\begin{theorem}
	When $\left\|\mathrm{A}\right\|_F \leq 1$, the convex envelope of $rank(\mathrm{A})$ is the nuclear-norm $\left\| \mathrm{A} \right\|_{\star}$.
\end{theorem}
The theorem is proved in~\cite{fazel2002matrix}. Our settings are slightly different from above theorem, we have $\left\|\mathrm{A}\right\|_F \leq \sqrt{B}$ as shown in Eqn.~\ref{tracemax}. Thus the convex envelope of $rank(\mathrm{A})$ becomes $ \left\| \mathrm{A} \right\|_{\star} / \sqrt{B}$, which is also proportional to $\left\| \mathrm{A} \right\|_{\star}$. Meanwhile, $rank(\mathrm{A})$ measures the diversity when $\left\| \mathrm{A} \right\|_F$ is near the upper-bound, as described by Eqn.~\ref{category} in Sec.~\ref{rank}. Therefore, if $\left\| \mathrm{A} \right\|_F$ is near $\sqrt{B}$, prediction diversity can be approximately represented by $\left\| \mathrm{A} \right\|_{\star}$. Therefore, maximizing $\left\| \mathrm{A} \right\|_{\star}$ ensures higher prediction diversity.

For matrix $\mathrm{A}$, nuclear-norm $\left\| \mathrm{A} \right\|_{\star}$ is defined as the sum of singular values of $\mathrm{A}$, calculated as follows:
\begin{equation}
	\begin{split}{
			\left\| \mathrm{A} \right\|_{\star} = \sum_{i=1}^{D} \sigma_i,
	}\end{split}
	\label{nnm}
\end{equation}
where $\sigma_i$ denotes the $i$th largest singular value. The number of the singular values is denoted as $D$, the smaller one in $B$ and $C$, {\it i.e.}, $D=\min(B,C)$. For batch prediction output matrix $\mathrm{A}$, $\left\| \mathrm{A} \right\|_{\star}$ is called batch nuclear-norm.

To explore the discriminability in $\left\| \mathrm{A} \right\|_{\star}$, we analyze the relationship between $\left\| \mathrm{A} \right\|_{\star}$ and $\left\| \mathrm{A} \right\|_F$. We find that $\left\| \mathrm{A} \right\|_F$ can be also expressed by singular values $\sigma_i$ follows:
\begin{equation}
	\begin{split}{
			\left\| \mathrm{A} \right\|_F = \sqrt{\sum_{i=1}^{D} \sigma_i^2},
	}\end{split}
	\label{Fnorm}
\end{equation}
where the calculation is shown in Sect. 1.2 in Supplementary in detail. With auxiliary singular values, the relationship between $\left\| \mathrm{A} \right\|_{\star}$ and $\left\| \mathrm{A} \right\|_F$ is demonstrated as follows:
\begin{equation}
	\begin{split}{
			\frac{1}{\sqrt{D}}\left\| \mathrm{A} \right\|_{\star} \leq \left\| \mathrm{A} \right\|_F \leq \left\| \mathrm{A} \right\|_{\star} \leq \sqrt{D} \cdot \left\| \mathrm{A} \right\|_F.
	}\end{split}
	\label{lep}
\end{equation}
Similar as~\cite{fazel2002matrix,recht2010guaranteed,srebro2005maximum}, $\left\| \mathrm{A} \right\|_{\star}$ and $\left\| \mathrm{A} \right\|_F$ could bound each other. Therefore, $\left\| \mathrm{A} \right\|_F$ tends to be larger, if $\left\| \mathrm{A} \right\|_{\star}$ becomes larger. Since maximizing $\left\| \mathrm{A} \right\|_F$ could increase the discriminability as described in Sec.~\ref{fnorm}, maximizing $\left\| \mathrm{A} \right\|_{\star}$ also contributes to the improvement on  discriminability.

Due to the relationship between $\left\| \mathrm{A} \right\|_{\star}$ and $\left\| \mathrm{A} \right\|_F$ in Eqn.~\ref{lep}, and the fact that upper-bound of $\left\| \mathrm{A} \right\|_F$ is $\sqrt{B}$ in Eqn.~\ref{tracemax}, we calculate the maximum of $\left\| \mathrm{A} \right\|_{\star}$ as follows:
\begin{equation}
	\begin{split}{
			\left\| \mathrm{A} \right\|_{\star}	\leq \sqrt{D} \cdot \left\| \mathrm{A} \right\|_F \leq \sqrt{D \cdot B},
	}\end{split}
	\label{lepa}
\end{equation}
where the two inequality conditions in the equation correspond to the two influence factors of $\left\| \mathrm{A} \right\|_{\star}$ respectively. The first inequality corresponds to the diversity, and the second to the discriminability. When prediction diversity is higher, the rank of $\mathrm{A}$ tends to be larger with increased $\left\| \mathrm{A} \right\|_{\star}$. Similarly, high prediction discriminability accompanies with increased $\left\| \mathrm{A} \right\|_F$ and large $\left\| \mathrm{A} \right\|_{\star}$.

\subsubsection{Fast Batch Nuclear-norm Calculation}
\label{fbnm-part}
 
%

To obtain the nuclear-norm, we are supposed to calculate all the singular values in the matrix $\mathrm{A}$. The SVD computed on the matrix $\mathrm{A} \in \mathbb{R}^{B \times C}$ costs $O(\min(B^2C, BC^2))$ time complexity. This computation complexity could be simply denoted as $O(n^3)$, where $n$ denotes the size scale of batch output matrix. In normal circumstance, $B$ and $C$ are small, and the overall computational budget of $\left\| \mathrm{A} \right\|_{\star}$ is almost negligible in the training procedure of deep networks. However, when $B$ and $C$ are large, the computation complexity is growing and the calculation of $\left\| \mathrm{A} \right\|_{\star}$ becomes time-consuming. Worse still, the calculation of singular value decomposition might not be converged in some cases, see the experiments in Table~\ref{SSDA}. So it is necessary to seek for approximations of the singular values towards fast and efficient calculation.

The singular values $\sigma_i$ are the main components of $\mathrm{A}$. For $\mathrm{A}_i$, only a few categories have non-zero responses. Thus $\mathrm{A}$ is sparse, and its singular values could be approximated by the combination of category responses in $\mathrm{A}$.
\begin{theorem}
If $\left\| \mathrm{A} \right\|_F$ is near the upper-bound $\sqrt{B}$, the $i$th largest singular value $\sigma_i$ can be approximated by:
\begin{equation}
	\sigma_j \approx top(\sum_{i=1}^B \mathrm{A}_{i,j}^2,j) \qquad \forall i \in {0,...,D}.
	\label{cls} \end{equation}
\end{theorem}
The theorem is proved in Sect. 1.4 in Supplementary. With approximation of the singular values, we calculate the batch nuclear-norm in a much faster way, as follows:
\begin{equation}
	\left\| \mathrm{\hat{A}} \right\|_{\star} =\sum_{j=1}^{D} top(\sqrt{\sum_{i=1}^B} \mathrm{A}_{i,j}^2,j).
	\label{aproximate} \end{equation}
This equation means that in terms of $\mathrm{A}$, the main components of the $L_{1,2}$-norm could approximate nuclear-norm, if $\left\| \mathrm{A} \right\|_F$ is near the upper-bound $\sqrt{B}$. While the rest of entries of the $L_{1,2}$-norm are regarded as noisy parts in $\mathrm{{A}}$.

Our fast batch nuclear-norm calculation maintains two advantages compared with SVD used in~\cite{cui2020nnm}. First, the computation complexity of $\left\| \mathrm{\hat{A}} \right\|_{\star}$ is $O(\min(BC, B^2))$, denoted as $O(n^2)$ for simplicity, which reduces the computation complexity from $O(n^3)$ to $O(n^2)$. Second,  our fast computation of $\left\| \mathrm{\hat{A}} \right\|_{\star}$ is based on ordinary floating point calculation of matrix components, without any risk of non-convergence.



\subsubsection{Extreme Points of Batch Nuclear-norm}

We have proved that maximizing $\left\| \mathrm{A} \right\|_{\star}$ attains improvement on both prediction discriminability and diversity. For better comprehension of the batch nuclear-norm, we construct a toy examples to demonstrate the properties of extreme points. 

We assume $B$ and $C$ are $2$. In this case, $\mathrm{A}$ could be expressed as:
\begin{equation}
	\mathrm{A}=\left[
	\begin{array}{ccc} 
		x &    1-x   \\ 
		y &    1-y  \\ 
	\end{array}
	\right],
\end{equation}
where $x$ and $y$ are variables. Thus the negative entropy, $F$-norm, nuclear-norm and fast nuclear-norm could be calculated as:
\begin{equation}
	\begin{split}{
			-H(\mathrm{A})=&x\log(x)+(1-x)\log(1-x)+y\log(y)\\
			&+(1-y)\log(1-y)\\
			\left\| \mathrm{A} \right\|_F =& \sqrt{x^2+(1-x)^2+y^2+(1-y)^2}\\
			\left\| \mathrm{A} \right\|_{\star} =& \sqrt{x^2+(1-x)^2+y^2+(1-y)^2+2|y-x|} \\
			\left\| \mathrm{\hat{A}} \right\|_{\star} =& \sqrt{x^2+(1-x)^2}+\sqrt{y^2+(1-y)^2},
	}\end{split}
\end{equation}
where the calculation of $\left\| \mathrm{A} \right\|_{\star}$ is described in Sect. 1.5 in Supplementary. For entropy and $F$-norm, there is no constraint on the relationship between $x$ and $y$, thus negative entropy and $F$-norm could reach the maximum when:
\begin{equation}
	\mathrm{A}=\left[
	\begin{array}{ccc} 
		1 &    0   \\ 
		1 &    0  \\ 
	\end{array}
	\right],
	\left[
	\begin{array}{ccc} 
		0 &    1   \\ 
		1 &    0  \\ 
	\end{array}
	\right],
	\left[
	\begin{array}{ccc} 
		1 &    0   \\ 
		0 &    1  \\ 
	\end{array}
	\right],
	\left[
	\begin{array}{ccc} 
		0 &    1   \\ 
		0 &    1  \\ 
	\end{array}
	\right].
\end{equation}
But $\left\| \mathrm{A} \right\|_{\star}$ and $\left\| \mathrm{\hat{A}} \right\|_{\star}$ will reach the maximum only when: 
\begin{equation}
	\mathrm{A}=
	\left[
	\begin{array}{ccc} 
		0 &    1   \\ 
		1 &    0  \\ 
	\end{array}
	\right],
	\left[
	\begin{array}{ccc} 
		1 &    0   \\ 
		0 &    1  \\ 
	\end{array}
	\right],
\end{equation}
where $\left\| \mathrm{A} \right\|_{\star}$ and $\left\| \mathrm{\hat{A}} \right\|_{\star}$ enforce diversity by maximizing the prediction divergences among the batch data. The maximum of $\left\| \mathrm{\hat{A}} \right\|_{\star}$ is calculated in Sect. 1.6 in Supplementary. Meanwhile, all the measurements will reach the minimum if and only if:
\begin{equation}
	\mathrm{A}=
	\left[
	\begin{array}{ccc} 
		0.5 &    0.5  \\ 
		0.5 &    0.5  \\ 
	\end{array}
	\right],
\end{equation}
which satisfies the situation with the lowest prediction discriminability and diversity.


\subsection{Batch Nuclear-norm for Domain Adaptation}
In domain adaptation, we are given source domain $\mathcal{D}_S$ and target domain $\mathcal{D}_T$. There are $N_S$ labeled source samples $\mathcal{D}_S=\{(x_i^S,y_i^S)\}_{i=1}^{N_S} $ in $C$ categories and $N_T$ unlabeled target samples $\mathcal{D}_T=\{x_i^T\}_{i=1}^{N_T}$ in the same $C$ categories. In $\mathcal{D}_S$, the labels are denoted as $y_i^S =[y_{i1}^S,y_{i2}^S,...,y_{iC}^S]\in \mathbb{R}^{C}$, where $y_{ij}^S$ equals to $1$ if $x_i^S$ belongs to the $j$th category and otherwise $0$. 

For object recognition, the classification responses are acquired by deep network $G$, {\it i.e.}, $\mathrm{A}_i=G(x_i)$. Typically network $G$ consists of a feature extraction network, a classifier and a softmax layer. With randomly sampled batch of $B_S$ samples $\{X^S,Y^S\}$ on the source domain, the classification loss on $\mathcal{D}_S$ could be calculated as:
\begin{equation}
	\mathcal{L}_{cls} =\frac{1}{B_S} \left \| Y^S{\log(G(X^S))}\right \|_1,
	\label{cls} \end{equation}
where $\left \| \cdot \right \|_1$ denotes the $L_1$-norm. Minimizing classification loss on source domain provides initial model parameters for further optimization.
\subsubsection{Batch Nuclear-norm Maximization and Minimization}
\begin{figure}[t]
	\begin{center}
		\includegraphics[width=0.98\linewidth]{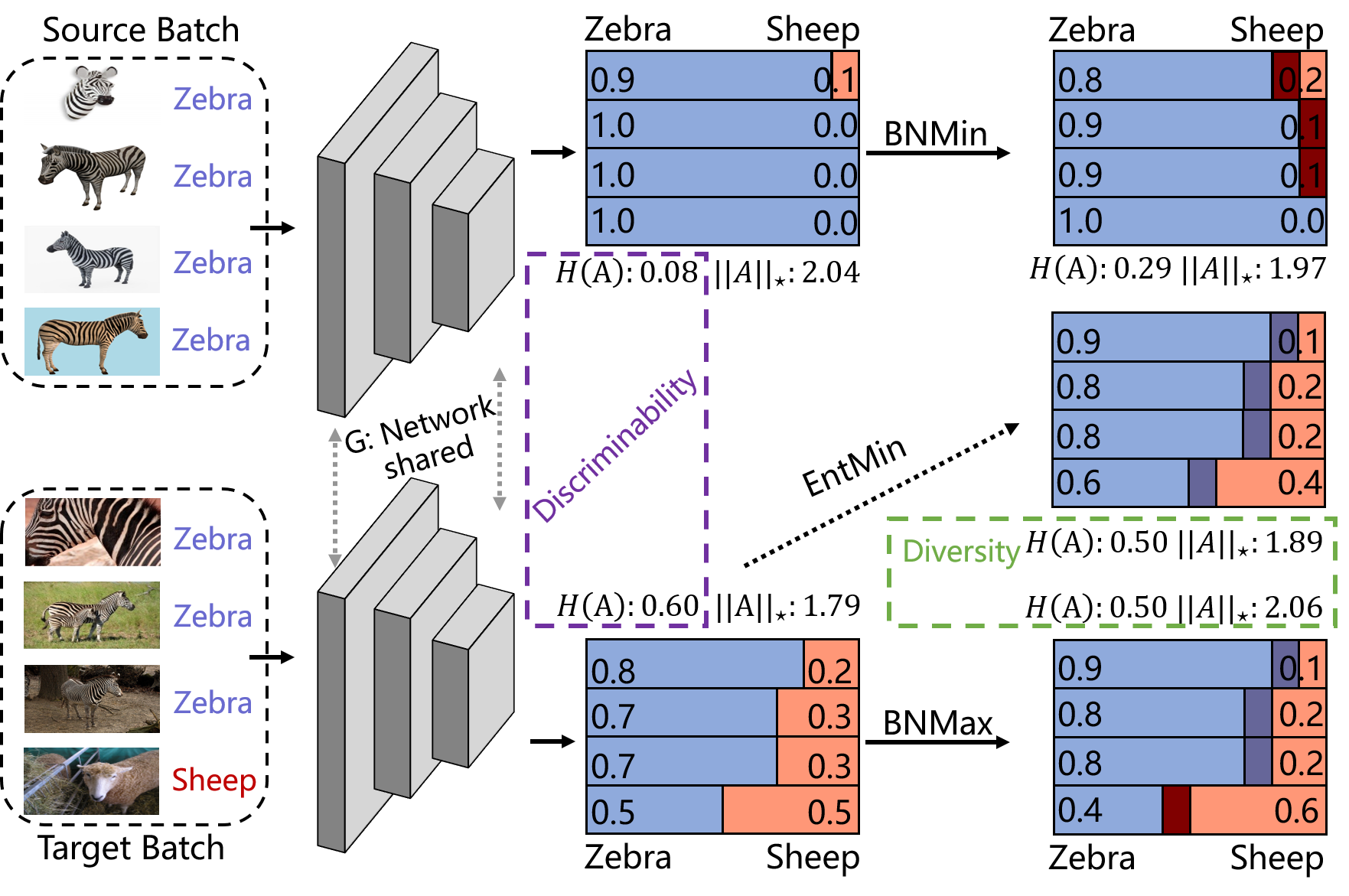}
	\end{center}
	\caption{Illustration of the Batch Nuclear-norm Maximization and Minimization (BNM$^2$), which is composed of BNMin and BNMax in a toy example of 4 batch size and 2 categories. The dark region means the variance of the variable, {\it i.e.}, the dark blue (red) represents the increase of blue (red) variable. $H(\mathrm{A})$ represents the entropy value and the value of nuclear-norm is denoted as $\left\| \mathrm{A} \right\|_{\star}$.}
	\label{softmatching}
\end{figure}

On target domain $\mathcal{D}_T$, on a randomly sampled batch of $B_T$ examples $\{X^T\}$, the classification response matrix is denoted as $G(X^T)$. To improve both the prediction discriminability and diversity, we propose Batch Nuclear-norm Maximization (BNMax) by maximizing the nuclear-norm of the batch matrix $G(X^T)$ on target domain. The loss function of BNMax, denoted as $\mathcal{L}_{BNMax}$ can be formulated as:
\begin{equation}
	\mathcal{L}_{BNMax}= -\frac{1}{B_T}\left \|G(X^T)\right \|_{\star},
	\label{opt}
\end{equation}
where the neural network $G$ is shared between $D_S$ and $D_T$. Minimizing $\mathcal{L}_{BNMax}$ reduces the target data density near the decision boundary without diversity degradation. With maintained prediction diversity, more determined predictions are gained, compared with entropy minimization.

On source domain $\mathcal{D}_S$, the classification response matrix of sampled batch $B_S$ examples $\{X^S\}$ is denoted as $G(X^S)$. The overly confident source predictions $G(X^S)$ indicates that $G(X^S)$ might be overfitting to source domain, resulting in difficulties for adapting to target domain. Thus to lessen the prediction discriminability, we propose Batch Nuclear-norm Minimization (BNMin) by minimizing nuclear-norm of $G(X^S)$ on source domain. The loss function of BNMin, denoted as $\mathcal{L}_{BNMin}$ can be calculated as:
\begin{equation}
	\mathcal{L}_{BNMin}= \frac{1}{B_S}\left \|G(X^S)\right \|_{\star}.
	\label{opt}
\end{equation}
Minimizing $\mathcal{L}_{BNMin}$ could reduce the distance from source data to the decision boundary, with less prediction discriminability on source domain. 

The approximation constraint of Eqn.~\ref{aproximate} is the combination of components in the matrix, where the gradient can be directly calculated. Meanwhile, the gradient of nuclear-norm could be calculated according to \cite{papadopoulo2000estimating}. Thus $\mathcal{L}_{BNMax}$, $\mathcal{L}_{BNMin}$ and the approximation constraints can be directly optimized by the gradient-based method in an end-to-end manner. To train the network, we simultaneously optimize classification loss $\mathcal{L}_{cls}$, BNMax loss $\mathcal{L}_{BNMax}$ and BNMin loss $\mathcal{L}_{BNMin}$, combined with the parameter $\lambda$ as:
\begin{equation}
	\begin{split}{
			\mathcal{L}_{BNM} &= \mathcal{L}_{cls} + \lambda \cdot (\mathcal{L}_{BNMax}), \\
			\mathcal{L}_{BNM^2} &= \mathcal{L}_{cls} + \lambda \cdot (\mathcal{L}_{BNMin} + \mathcal{L}_{BNMax}).
	}\end{split}
	\label{opt}
\end{equation}
We denote Batch Nuclear-norm Maximization as BNM ($\mathcal{L}_{BNM}$), and Batch Nuclear-norm Maximization and Minimization as BNM$^2$ ($\mathcal{L}_{BNM^2}$). When Batch Nuclear-Norm is replaced by Fast Batch Nuclear-norm, the loss functions with Eqn.~\ref{nnm} are replaced by Eqn.~\ref{aproximate}. Then the corresponding loss functions calculated in the fast manner are respectively denoted as FBNM and FBNM$^2$.

The key insight of BNM is sacrificing a certain level of the target prediction hit-rate on majority categories, to enhance the target prediction hit-rate on minority categories. 
With the increased diversity, the samples belonging to the majority classes might be misclassified as minority classes. Fortunately, the classification loss $\mathcal{L}_{cls}$ on source domain would penalize the wrongly encouraged diversity for target predictions, since model $G$ is shared between source and target domain. Asymptotically, model $G$ tends to produce more diverse target predictions, given that the samples can be correctly predicted. 
Competing against the situation that the majority categories tend to dominant the predictions of target batch samples, BNMax is effective to avoid mode collapse and prediction degradation for domain adaptation under both balanced and imbalanced category distributions. 
Meanwhile, BNMin prevents overfitting on source data, so the source domain knowledge tends to be more generalized and transferable.

To provide comprehensive understanding of our methods, we show a toy example in Figure~\ref{softmatching}. We assume that there are only two categories, {\it i.e.}, zebra and sheep. There are $4$ zebras in the source batch, and $3$ zebras and a goat in the target batch, so zebra is the majority category. Before domain adaptation, there exists large discriminability discrepancy between the $H(\mathrm{A})$, {\it i.e.}, $0.08$ for source batch and $0.60$ for target batch. 
This discrepancy is mitigated by both BNMin on source batch, and BNMax on target batch. As the key role in our method, BNMax ensures both the prediction discriminability and diversity for target domain. Higher prediction discriminability is represented by higher $H(\mathrm{A})$ after domain adaptation. However, with the same $H(\mathrm{A})$, the improvement on diversity in the target domain is shown by $\left\| \mathrm{A} \right\|_{\star}$ with BNMax and Entropy minimization (EntMin).


\subsubsection{Multi-batch Nuclear-norm Optimization}
In some cases, the number of categories $C$ is quite large, then batch size $B$ should not be too small during the training procedure. However, large $B$ and $C$ might lead to the difficulty on calculating $\mathrm{A}$, due to the limited memory size. Thus, it is necessary seek for efficient calculation under the setting of large $C$ and small $B$.
Specifically, we record $K$ previous prediction matrices of $\mathrm{A}$ and concatenate these matrices into a large matrix $\mathrm{R} \in \mathbb{R}^{(K\times B)\times C}$. With enough $K$ records, we calculate BNMax and BNMin loss functions based on $\mathrm{R}$. Among the $K$ matrices, only the $K$th matrix propagates the gradient of the loss functions. After the gradient propagation, the $K$ matrices are released for recording the next $K$ prediction matrices into $\mathrm{R}$. The optimization process of BNM$^2$ with matrix $\mathrm{R}$ is shown in Algorithm~\ref{step2}. When $K$ is large enough, matrix $\mathrm{R}$ could contain predictions for the whole dataset. When $K$ is $1$, the optimization process reduces to the single batch setting. We set $K=3$ in our experiments for ensuring good accuracy and balanced computation.
Line 5 in Algorithm can also be replaced with the approximated objective functions of FBNM and FBNM$^2$ using Eqn.~\ref{aproximate}.

\begin{algorithm}[t]
	\caption{Multi-batch Nuclear-norm for BNM$^2$} 
	
	\begin{algorithmic}[1]
		 \REQUIRE network $G$, selected source data $X^S_i$ and target data $X^T_i$ for the $i$th iteration, with total $M$ iterations and $K$. 
		\ENSURE  network weights of $G$.
		\STATE $\mathrm{R}^S$, $\mathrm{R}^T$ = [\ ], [\ ]
		\FOR{$i=1$ to $M$}
		\STATE $\mathcal{L} = \frac{1}{B_S} \left \| Y^S{\log(G(X^S))}\right \|_1$
		\IF { $\mod(i,K)==0$
		}
		\STATE $\mathcal{L} \quad += \frac{1}{B_S}\left \|[G(X^S);R^S]\right \|_{\star} -\frac{1}{B_T}\left \|[G(X^T);R^T]\right \|_{\star}$
		\STATE $R^S$, $R^T$ = [\ ], [\ ]
		\ELSE
		\STATE $\mathrm{R}^S = [G(X^S);\mathrm{R}^S]$
		\STATE $\mathrm{R}^T = [G(X^T);\mathrm{R}^T]$
	    \ENDIF
	    \STATE Update $G$.
	    \ENDFOR

	\end{algorithmic}
	\label{step2}
\end{algorithm}

\section{Experiments}
We apply our method to three domain adaptation tasks, {\it i.e.}, unsupervised domain adaptation (UDA), semi-supervised domain adaptation (SSDA), and unsupervised open domain recognition (UODR). The experiments of the three tasks are accomplished on datasets of Office-31 \cite{saenko2010adapting}, Office-Home \cite{venkateswara2017deep}, Balanced Domainnet, Semi Domainnet~\cite{saito2019semi} and I2AwA~\cite{zhuo2019unsupervised}. We also denote the direct entropy minimization and batch Frobenius-norm maximization as EntMin and BFM in our experiments. When BNM and other methods cooperate with existing methods, we concatenate the names with ``+".

\subsection{Unsupervised Domain Adaptation}

Office-31 \cite{saenko2010adapting} and Office-Home \cite{venkateswara2017deep} are standard benchmarks for unsupervised domain adaptation. Office-31 contains 4,110 images in 31 categories, and consists of three domains: Amazon (A), Webcam (W), and DSLR (D). We evaluate the methods across the three domains, resulting in six transfer scenarios. Office-Home is a relative challenging dataset with 15,500 images in 65 categories. It has four significantly different domains: Artistic images (Ar), Clip Art (Cl), Product images (Pr), and Real-World images (Rw). There are 12 transfer tasks among four domains in total.

\begin{table}[t]
	\begin{center}
		
		
		\caption{Accuracies (\%) on Office-31 for ResNet50-based unsupervised domain adaptation methods.}
		\vspace{-10pt}
		\label{tableoffice}
			\setlength{\tabcolsep}{0.3mm}{ 
				\resizebox{0.48\textwidth}{!}{%
				\begin{tabular}{cccccccc}
					\hline
					Method & A$\rightarrow$D & A$\rightarrow$W & D$\rightarrow$W & W$\rightarrow$D & D$\rightarrow$A & W$\rightarrow$A & Avg \\
					\hline\hline
					ResNet-50 \cite{he2016deep} & 68.9 & 68.4 & 96.7 & 99.3 & 62.5 & 60.7 & 76.1 \\
					GFK \cite{gong2012geodesic} & 74.5 & 72.8 & 95.0 & 98.2 & 63.4 & 61.0 & 77.5 \\
					DAN \cite{long2015learning} & 78.6 & 80.5 & 97.1 & 99.6 & 63.6 & 62.8 & 80.4 \\
					DANN \cite{ganin2016domain} & 79.7 & 82.0 & 96.9 & 99.1 & 68.2 & 67.4 & 82.2 \\
					ADDA \cite{tzeng2017adversarial} & 77.8 & 86.2 & 96.2 & 98.4 & 69.5 & 68.9 & 82.9 \\
					MaxSquare \cite{chen2019domain}&90.0 &92.4 &\textbf{99.1} &\textbf{100.0}  &68.1 &64.2 &85.6\\
					Simnet \cite{pinheiro2018unsupervised} & 85.3 & 88.6 & 98.2 & 99.7 & {73.4} & 71.8 & 86.2 \\				
					GTA \cite{sankaranarayanan2018generate} & 87.7 & 89.5 & 97.9 & 99.8 & 72.8 & 71.4 & 86.5 \\
					MCD \cite{saito2018maximum} & {92.2} & {88.6} & {98.5} & \textbf{100.0} & {69.5} & {69.7} & {86.5} \\	
					CBST~\cite{zou2018unsupervised}  &86.5  &87.8  &98.5 &\textbf{100.0}  &70.9  &71.2  &85.8 \\
					CRST~\cite{zou2019confidence}   &88.7  &89.4  &98.9 &\textbf{100.0} &70.9  &72.6 &86.8\\

					\hline
					EntMin  &86.0	&87.9	&98.4	&\textbf{100.0} 	&67.0		&63.7&83.8\\
					BFM &87.7	&86.9	&98.5	&\textbf{100.0}	&67.6	&63.0	&84.0 \\
					
					BNM  & {90.3} & {91.5} & {98.5} & \textbf{100.0} & {70.9}  & {71.6} & {87.1} \\
					FBNM &90.4 	&91.9 	&98.6 	&99.9 	&71.2 	&70.5 	&87.1\\ 
					BNM$^2$ &92.6 	&92.6 	&98.7 	&99.9 	&72.4 	&73.6 	&88.3 \\
					FBNM$^2$ &91.6 &92.7 &98.5 &100.0 	&72.1 &73.1 &88.0 \\				
					\hline
					CDAN \cite{long2018conditional} & {92.9} & \textbf{93.1} & {98.6} & \textbf{100.0} & {71.0} & {69.3} & {87.5} \\
					CDAN+EntMin &92.0	&91.2 &98.7	&\textbf{100.0}	&70.7		&71.0	&87.3\\
					CDAN+BNM  & {92.9} & {92.8} & {98.8} & \textbf{100.0} & {73.5} & {73.8} & {88.6} \\
					CDAN+FBNM & 92.7 	&92.4 		&98.6 &99.7 	&74.0 &73.7 		&88.5 \\
					CDAN+BNM$^2$ & {93.4} 		&92.7 		&99.0 &100.0 	&73.5 &75.5 	&{89.0}\\
					CDAN+FBNM$^2$ &92.6 		&92.4 			&99.0 &100.0 &{74.7} &{74.2} 	&{88.8} \\
					\hline
					SHOT~\cite{liang2020we}  &\textbf{94.0} &90.1  &98.4  &99.9 &74.7 &74.3&88.6 \\
					BNM-S &93.0 	&93.0  &	98.2	&99.9  	&\textbf{75.4}	&74.9 	&\textbf{89.1} \\
					FBNM-S &93.0 	&92.9  &98.2  &99.9 &\textbf{75.4}		&\textbf{75.0} &\textbf{89.1}\\
					\hline
			\end{tabular}}
			
		}
	\end{center}
	\vspace{-10pt}
\end{table}

\begin{table*}[htbp]
	\begin{center}
		
		\caption{Accuracies (\%) on Office-Home for ResNet50-based unsupervised domain adaptation methods.}
		
		\vspace{-5pt}
		\label{tableofficehome}
		\addtolength{\tabcolsep}{-5.5pt}
		\scalebox{0.95}{
			\resizebox{\textwidth}{!}{%
				\begin{tabular}{cccccccccccccc}
					\hline
					Method& Ar$\rightarrow$Cl & Ar$\rightarrow$Pr & Ar$\rightarrow$Rw & Cl$\rightarrow$Ar & Cl$\rightarrow$Pr & Cl$\rightarrow$Rw & Pr$\rightarrow$Ar & Pr$\rightarrow$Cl & Pr$\rightarrow$Rw & Rw$\rightarrow$Ar & Rw$\rightarrow$Cl & Rw$\rightarrow$Pr & Avg \\
					\hline\hline
					
					ResNet-50 \cite{he2016deep} & 34.9 & 50.0 & 58.0 & 37.4 & 41.9 & 46.2 & 38.5 & 31.2 & 60.4 & 53.9 & 41.2 & 59.9 & 46.1 \\
					DAN \cite{long2015learning} & 43.6 & 57.0 & 67.9 & 45.8 & 56.5 & 60.4 & 44.0 & 43.6 & 67.7 & 63.1 & 51.5 & 74.3 & 56.3 \\
					DANN \cite{ganin2016domain} & 45.6 & 59.3 & 70.1 & 47.0 & 58.5 & 60.9 & 46.1 & 43.7 & 68.5 & 63.2 & 51.8 & 76.8 & 57.6 \\
					MCD \cite{saito2018maximum} &48.9	&68.3	&74.6	&61.3	&67.6	&68.8	&57	&47.1	&75.1	&69.1	&52.2	&79.6	&64.1 \\
					
					SAFN~\cite{Xu_2019_ICCV} &52.0&	71.7&76.3&64.2&69.9&71.9&63.7&51.4&	77.1&70.9&57.1&81.5&67.3\\
					Symnets \cite{zhang2019domain} & 47.7 & 72.9 & 78.5 & {64.2} & 71.3 & 74.2 & {64.2}  & 48.8 & 79.5 & {74.5} & 52.6 & {82.7} & 67.6 \\
					MDD \cite{zhang2019bridging} & 54.9 & 73.7 & 77.8 & 60.0 & 71.4 & 71.8 & 61.2 & 53.6 & 78.1 & 72.5 & \textbf{60.2} & 82.3 & 68.1 \\
					\hline
					
					EntMin  &43.2	&68.4	&78.4	&61.4	&69.9	&71.4	&58.5	&44.2	&78.2	&71.1	&47.6	&81.8	&64.5 \\
					BFM &43.3 	&69.1 	&78.0 	&61.3 	&67.4 	&70.9 	&57.8 	&44.1 	&78.9 	&72.1 	&50.1 	&81.0 	&64.5 \\
					BNM &{52.3}	&{73.9}	&{80.0}	&63.3	&{72.9}	&{74.9}	&61.7	&{49.5}	&{79.7}	&70.5	&{53.6}	&82.2	&67.9\\	
					FBNM &54.9 	&73.4 	&79.3 	&62.6 	&73.6 	&74.5 	&61.0 	&51.7 	&79.6 	&70.1 	&57.1 	&82.3 	&68.3\\
					BNM$^2$ &55.8 	&74.8 	&79.9 	&64.2 	&73.7 	&75.5 	&61.6 	&52.4 	&80.4 	&72.2 	&56.8 	&83.5 	&69.2\\ 
					FBNM$^2$ &55.3 	&75.4 	&79.9 	&64.0 	&74.2 	&75.5 	&62.2 	&52.0 	&80.4 	&72.3 	&57.0 	&83.4 	&69.3\\ 
					\hline
					CDAN \cite{long2018conditional} & 50.7 & 70.6 & 76.0 & 57.6 & 70.0 & 70.0 & 57.4 & 50.9 & 77.3 & 70.9 & 56.7 & {81.6} & 65.8 \\
					CDAN+EntMin &54.1	&72.4	&78.3	&61.8	&71.8	&73.0	&62.0	&52.3	&79.7	&72.0	&57.0	&83.2	&68.1 \\

					CDAN+BNM &56.1 	&75.0 	&79.0 	&63.3 	&73.1 	&74.0 	&62.4 	&53.6 	&80.9 	&72.2 	&58.6 	&83.5 	&69.3 \\
					CDAN+BNM$^2$ &55.8 	&75.1 	&79.1 	&63.3 	&73.2 	&74.1 	&62.6 	&53.3 	&80.7 	&72.2 	&59.0 	&83.8 	&69.4\\
					CDAN+FBNM &56.3 	&74.6 	&79.3 	&63.4 	&73.2 	&73.9 	&62.7 	&53.5 	&81.0 	&72.7 	&58.7 	&83.6 	&69.4\\
					CDAN+FBNM$^2$ &56.1 	&74.9 	&79.3 	&63.6 	&73.6 	&74.1 	&62.9 	&54.1 	&81.0 	&72.5 	&58.4 	&83.6 	&69.5\\
					\hline 
					HDAN~\cite{cui2020hda} &56.8 &75.2 &79.8 &65.1 &73.9 &75.2 &66.3 &56.7 &81.8 &\textbf{75.4} &{59.7} &\textbf{84.7} &70.9\\
					HDAN+BNM &57.4 	&75.8 	&{80.3} 	&66.1 	&74.3 	&75.4 	&67.0 	&57.6 	&81.9 	&74.9 	&{59.7} 	&\textbf{84.7} 	&71.3 \\
					HDAN+BNM$^2$ &57.3 	&75.8 	&{80.3} 	&66.1 	&{74.7} 	&{75.8} 	&66.7 	&\textbf{58.0} 	&81.9 	&75.1 	&59.9 	&\textbf{84.7} 	&{71.4} \\
					HDAN+FBNM &\textbf{57.6} 	&{75.9} 	&{80.3} 	&65.7 	&74.1 	&{75.8} 	&{66.8} 	&57.7 	&{82.0} 	&74.9 	&59.6 	&\textbf{84.7} 	&71.3 \\ 
					HDAN+FBNM$^2$ &57.4 	&75.7 	&80.2 	&{66.3} 	&74.5 	&75.7 	&67.5 	&57.9 	&81.9 	&75.2 	&{59.7} 	&\textbf{84.7} 	&{71.4} \\
					\hline
					SHOT~\cite{liang2020we} &57.1 &\textbf{78.1} &81.5 &68.0 &\textbf{78.2} &78.1 &67.4 &54.9 &82.2 &73.3 &58.8 &84.3 &71.8 \\
					
					FBNM-S &56.9 	&77.7 	&81.6 	&\textbf{68.1} 	&77.6 	&79.2 	&\textbf{67.7} 	&55.3 	&\textbf{82.2} 	&73.7 	&59.8 	&\textbf{84.7} 	&{72.0} \\
					BNM-S &{57.4} 	&77.8 	&\textbf{81.7} 	&67.8 	&77.6 	&\textbf{79.3} 	&67.6 	&55.7 	&82.2 	&73.5 	&59.5 	&\textbf{84.7} 	&\textbf{72.1}\\
					\hline
		\end{tabular}}}
		\vspace{-10pt}
	\end{center}
\end{table*}

We adopt ResNet-50 \cite{he2016deep} pre-trained on ImageNet \cite{deng2009imagenet} as our backbone. The batch size is fixed to be $36$ in our experiments, implemented with PyTorch~\cite{torch}. During the training of the network, we employ mini-batch stochastic gradient descent (SGD) with initial learning rate as 0.0001 and momentum of 0.9 to train our model. When BNM is regarded as a typical method, the parameter $\lambda$ is fixed to be $0.5$. When BNM is combined with existing methods, the parameter $\lambda$ is set to be $0.25$. For each method, we run four random trials and report the average accuracy.

\begin{figure}[t]
	\begin{center}
		
		\begin{minipage}{.24\textwidth}
			\subfigure[Target]{
				\includegraphics[width=0.9\textwidth]{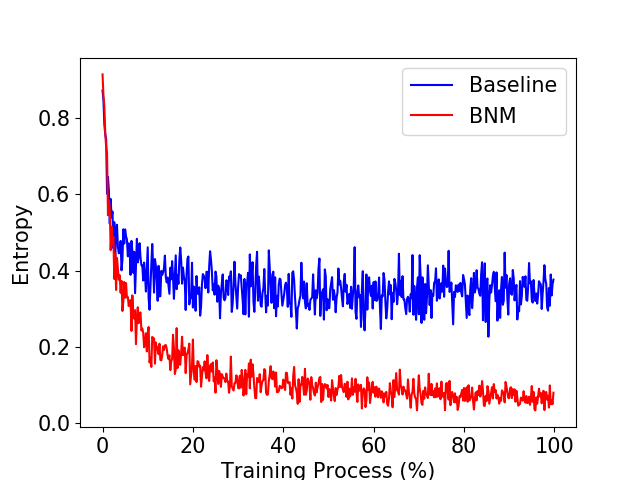}
				\label{entropytarget}}
		\end{minipage}
		\begin{minipage}{.24\textwidth}
			\subfigure[Source]{	
				\includegraphics[width=0.9\textwidth]{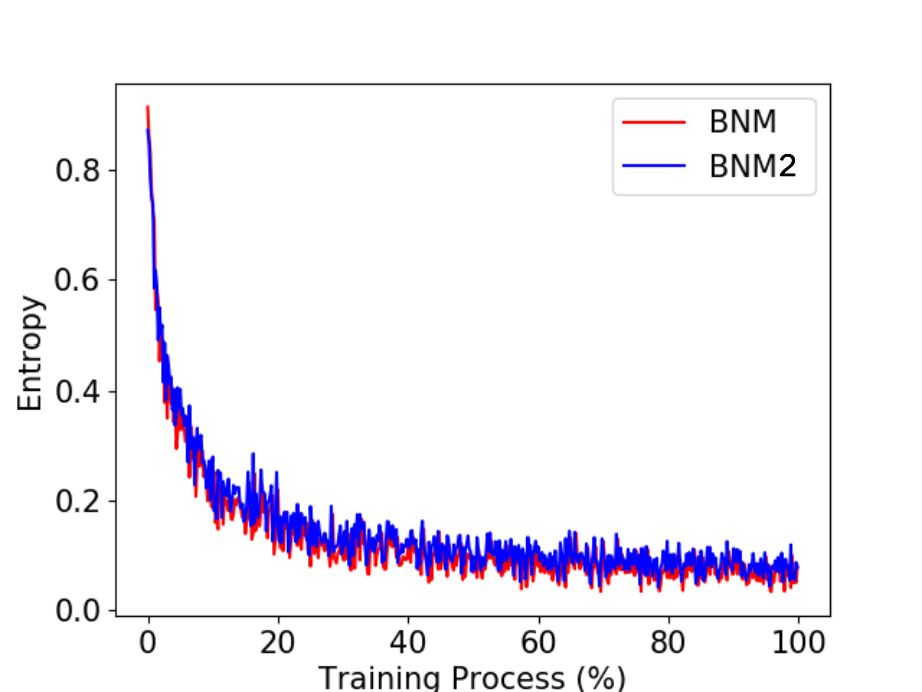}
				\label{entropysource}}
		\end{minipage}
	    \vspace{-3ex}
		\caption{Prediction discriminability for Ar $\rightarrow$ Cl. As a typical measurement, lower entropy means more prediction discriminability. }
		\label{entropy}
	\end{center}
	\vspace{-3ex}
\end{figure}

\begin{figure}[t]
	\begin{center}
		\begin{minipage}{.22\textwidth}
			\subfigure[Ar $\rightarrow$ Cl]{	
				\includegraphics[width=0.9\textwidth]{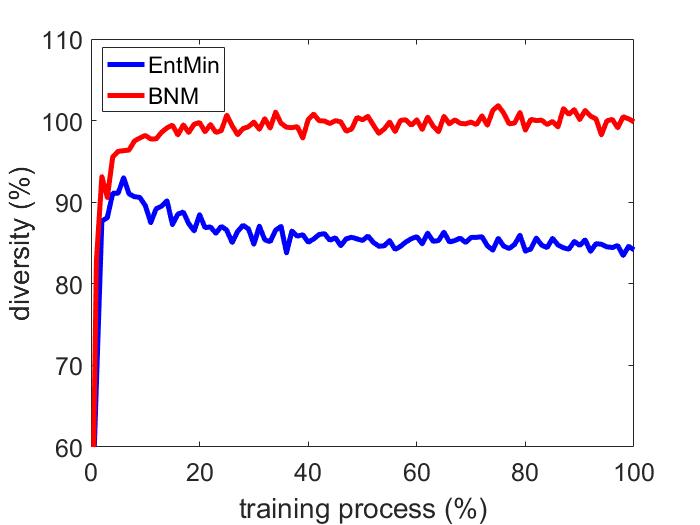}
				\label{home1}}
		\end{minipage}
		\begin{minipage}{.22\textwidth}
			\subfigure[Ar $\rightarrow$ Pr]{
				\includegraphics[width=0.9\textwidth]{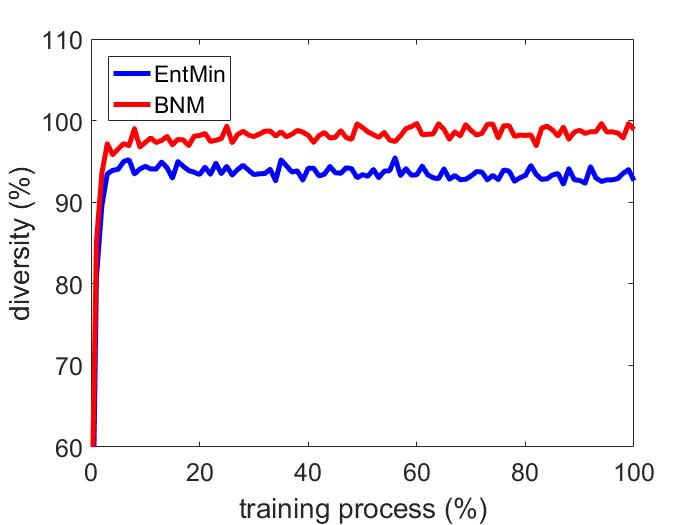}
				\label{home2}}
		\end{minipage}
		\vspace{-3ex}
		\caption{Diversity ratio on Office-Home for domain adaptation, calculated as the predicted diversity divided by the ground-truth diversity. The predicted/ground-truth diversity is measured by the average number of predicted/ground-truth categories in randomly sampled batches.}
		\label{diversity}
	\end{center}
	\vspace{-3ex}
\end{figure}

The results on Office-31 and Office-Home are shown in Table~\ref{tableoffice} and~\ref{tableofficehome}. On both Office-31 and Office-Home, as expected, BFM obtains similar results with EntMin, while BNM achieves substantial improvement on average over other entropy-based methods. Surprisingly, BNM obtains superior results compared with popular alignment-based competitors. The results show that Batch Nuclear-norm Maximization is effective for domain adaptation, especially on the difficult tasks where the baseline accuracy is relatively low, such as D$\rightarrow$A in Office-31 and Rw$\rightarrow$Cl in Office-Home. Beyond typical BNM, we also implement BNM$^2$, and it achieves even better results than BNM. Meanwhile, the approximated method FBNM achieves similar results as BNM. FBNM$^2$ further achieves similar results as BNM$^2$. The results show that our fast approximation methods are suitable to replace the original BNMax, BNMin or both.
\begin{table*}[t]
	\begin{center}
		\vspace{-5pt}
		\caption{ Accuracies (\%) on Balanced Domainnet of ResNet50-based UDA methods.}
		
		\label{balancedomainnet}
		\vspace{-5mm}
		\resizebox{\textwidth}{!}{%
			\begin{threeparttable}
				\centering
				\setlength{\tabcolsep}{1.3mm}{
					\begin{tabular}{ccccccccccccccccc}
						\hline 
						
						\multirow{2}{*}{Methods} & \multicolumn{2}{c}{R$\rightarrow$C} & \multicolumn{2}{c}{R$\rightarrow$P} & \multicolumn{2}{c}{P$\rightarrow$C} & \multicolumn{2}{c}{C$\rightarrow$S} & \multicolumn{2}{c}{S$\rightarrow$P} & \multicolumn{2}{c}{R$\rightarrow$S} & \multicolumn{2}{c}{P$\rightarrow$R} & \multicolumn{2}{c}{Avg}\\
						
						&{Src} &{Tgt} &{Src} &{Tgt} &{Src} &{Tgt} &{Src} &{Tgt} &{Src} &{Tgt} &{Src} &{Tgt} &{Src} &{Tgt} &{Src} &{Tgt} \\
						
						\hline \hline
						
						\multicolumn{1}{c}{Baseline1 \cite{he2016deep}} &\textbf{84.1} 	&67.3 	&84.1 	&70.6 	&\textbf{73.1} 	&59.5 	&\textbf{77.5} 	&62.0 	&71.7 	&58.5 	&\textbf{84.1} 	&64.6 	&\textbf{73.1} 	&71.7 	&\textbf{78.3} 	&64.9 \\ 
						
						\multicolumn{1}{c}{Baseline2 \cite{he2016deep}} &82.8 	&68.1 	&84.3 	&71.9 	&68.9 	&60.7 	&75.5 	&63.0 	&71.3 	&62.7 	&81.8 	&64.9 	&72.5 	&72.6 	&76.7 	&66.3 	  \\
						
						\multicolumn{1}{c}{DAN} &82.1 	&68.3 	&84.0 	&71.8 	&69.4 	&61.4 	&75.8 	&63.8 	&71.7 	&63.5 	&81.3 	&65.1 	&71.6 	&71.9 	&76.6 	&66.6  \\
						\multicolumn{1}{c}{DANN}&81.2 	&69.3 	&82.6 	&72.7 	&65.0 	&58.5 	&72.5 	&64.1 	&70.1 	&61.9 	&80.5 	&67.8 	&69.0 	&70.2 	&74.4 	&66.4 \\

						\multicolumn{1}{c}{ENT~\cite{grandvalet2005semi}} &82.0 &69.1 	&83.9 	&71.4 	&69.2 	&62.4 	&74.9 	&62.4 	&71.5 	&64.0 	&81.4 	&65.8 	&71.4 	&73.5 	&76.3 	&66.9  \\
						
						\multicolumn{1}{c}{SAFN~\cite{Xu_2019_ICCV}} &82.5 	&67.9 	&84.4 	&71.5 	&69.3 	&60.5 	&75.9 	&63.1 	&71.0 	&63.1 	&81.9 	&64.6 	&72.0 	&72.4 	&76.7 	&66.2  \\
						
						\multicolumn{1}{c}{CDAN~\cite{long2018conditional}} &80.8 	&70.2 	&82.1 	&73.0 	&66.9 	&61.8 	&72.9 	&65.8 	&70.0 	&63.2 	&80.2 	&69.3 	&69.7 	&71.1 	&74.7 	&67.8  \\
						\hline
						\multicolumn{1}{c}{BFM} &82.5 	&69.4 	&84.6 	&73.5 	&69.0 	&62.2 	&76.1 	&65.2 	&72.1 	&64.5 	&82.2 	&67.8 	&72.9 	&74.2 	&77.1 	&68.1\\
						\multicolumn{1}{c}{BNM($K=1$)} &82.5 &	69.7 	&84.6 	&73.8 	&68.5 	&62.6 	&76.6 	&65.6 	&72.4 	&65.5 	&81.9 	&67.8 	&73.0 	&74.4 	&77.1 	&68.5  \\
						\multicolumn{1}{c}{BNM($K=2$)} &82.4 	&70.2 	&84.4 	&74.2 	&69.2 	&63.8 	&76.0 	&65.9 	&72.2 	&65.7 	&81.4 	&67.9 	&73.1 	&74.6 	&77.0 	&68.9\\
						\multicolumn{1}{c}{BNM($K=3$)} &82.4 	&70.5 	&84.4 	&74.2 	&69.3 	&63.4 	&75.9 	&65.8 	&72.2 	&65.6 	&81.3 	&68.1 	&73.1 	&74.7 	&77.0 	&68.9\\
						\multicolumn{1}{c}{BNM$^2$} &82.2 	&70.4 	&83.9 	&73.8 	&68.5 	&63.4 	&76.5 	&65.9 	&71.8 	&65.5 	&81.3 	&67.7 	&71.9 	&74.3 	&76.6 	&68.7  \\
						\multicolumn{1}{c}{FBNM} &82.1 	&70.4 	&84.4 	&74.0 	&68.7 	&63.2 	&76.3 	&66.2 	&71.5 	&65.3 	&81.5 	&68.0 	&72.9 	&74.6 	&76.8 	&68.8  \\
						\multicolumn{1}{c}{FBNM$^2$} &82.1 &70.4 	&84.6 	&74.2 	&68.7 	&63.4 	&75.6 	&65.9 	&72.2 	&65.5 	&81.2 	&67.9 	&72.9 	&74.6 	&76.7 	&68.8  \\ 
						\hline
						\multicolumn{1}{c}{HDAN~\cite{cui2020hda}} &82.8 	&72.8 	&84.3 	&74.8 	&70.5 	&65.8 	&75.1 	&68.0 	&71.7 	&66.6 	&82.8 	&\textbf{71.6} 	&{73.0} 	&75.4 	&77.2 	&70.7  \\
						\multicolumn{1}{c}{HDAN+BNM} &82.3 	&72.7 	&84.3 	&74.7 	&70.5 	&\textbf{66.1} 	&76.3 	&68.0 	&\textbf{72.6} 	&\textbf{67.7} 	&82.5 	&71.3 	&72.7 	&75.5 	&77.3 	&\textbf{70.9}  \\
						\multicolumn{1}{c}{HDAN+BNM$^2$} &82.2 	&72.9 	&84.2 	&75.2 	&69.4 	&65.1 	&75.3 	&67.7 	&71.7 	&66.8 	&82.0 	&71.2 	&72.5 	&75.1 	&76.8 	&70.6 \\ 
						\multicolumn{1}{c}{HDAN+FBNM} &82.6 	&\textbf{73.2}	&\textbf{84.5} 	&\textbf{75.4} 	&69.4 	&65.1 	&76.0 	&\textbf{68.5} 	&72.0 	&67.0 	&82.3 	&\textbf{71.6} 	&72.7 	&\textbf{75.6} 	&77.1 	&\textbf{70.9} \\
						\multicolumn{1}{c}{HDAN+FBNM$^2$} &82.4 	&73.0 	&84.3 	&75.0 	&70.0 	&65.8 	&75.5 	&68.2 	&71.2 	&66.5 	&82.1 	&71.4 	&72.8 	&\textbf{75.6} 	&76.9 	&70.8\\
						
						\hline
				\end{tabular}}
				\renewcommand{\labelitemi}{}
				\vspace{-5pt}
			\end{threeparttable}
		}
	\end{center}
\end{table*}
\begin{figure*}
	\begin{center}
		\begin{minipage}{.98\textwidth}
			\begin{minipage}{.24\textwidth}
				\subfigure[Baseline]{	
					\includegraphics[width=0.9\textwidth]{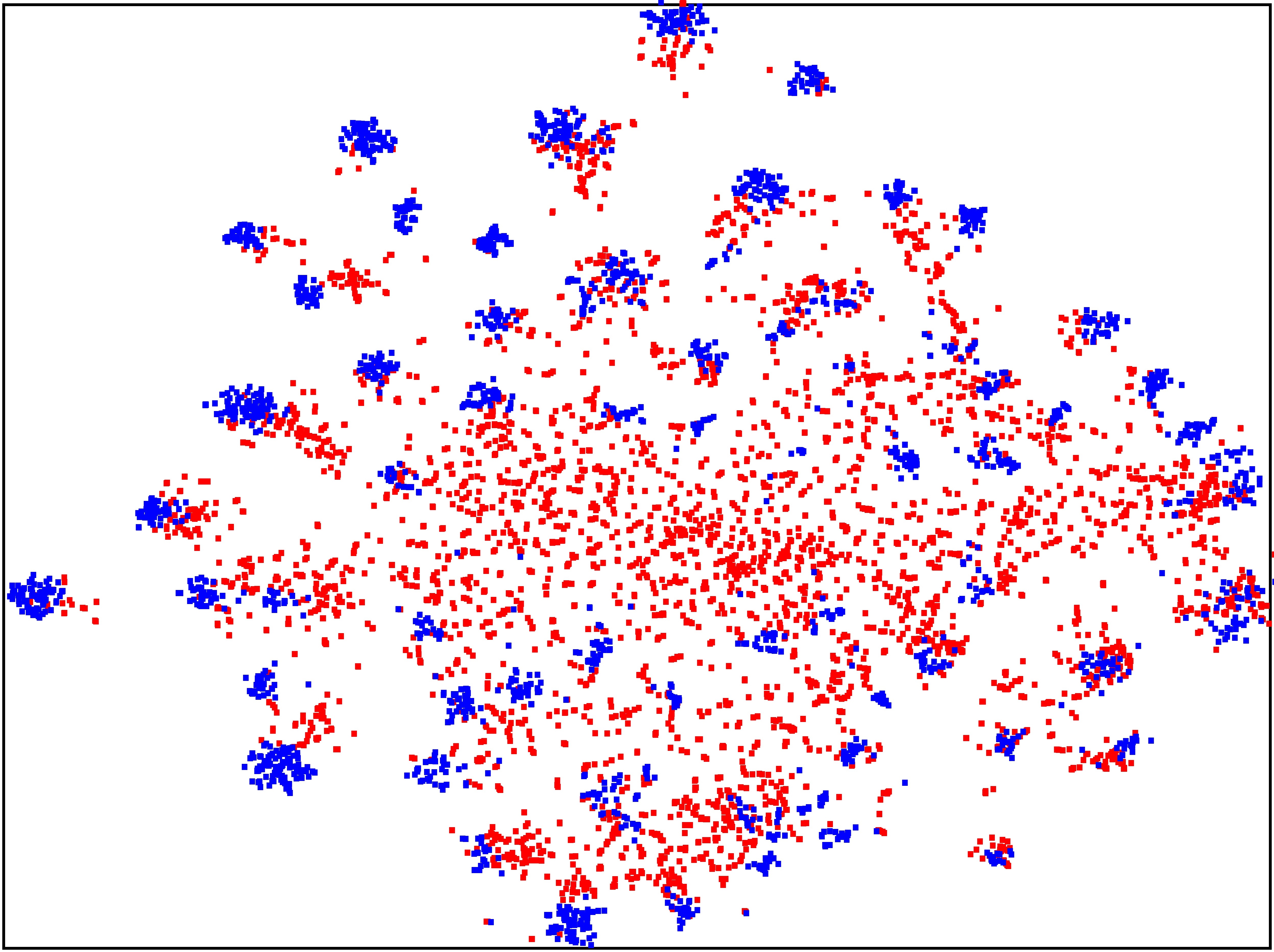}
					\label{tsne0}}
			\end{minipage}
			\begin{minipage}{.24\textwidth}
				\subfigure[BFM]{
					\includegraphics[width=0.9\textwidth]{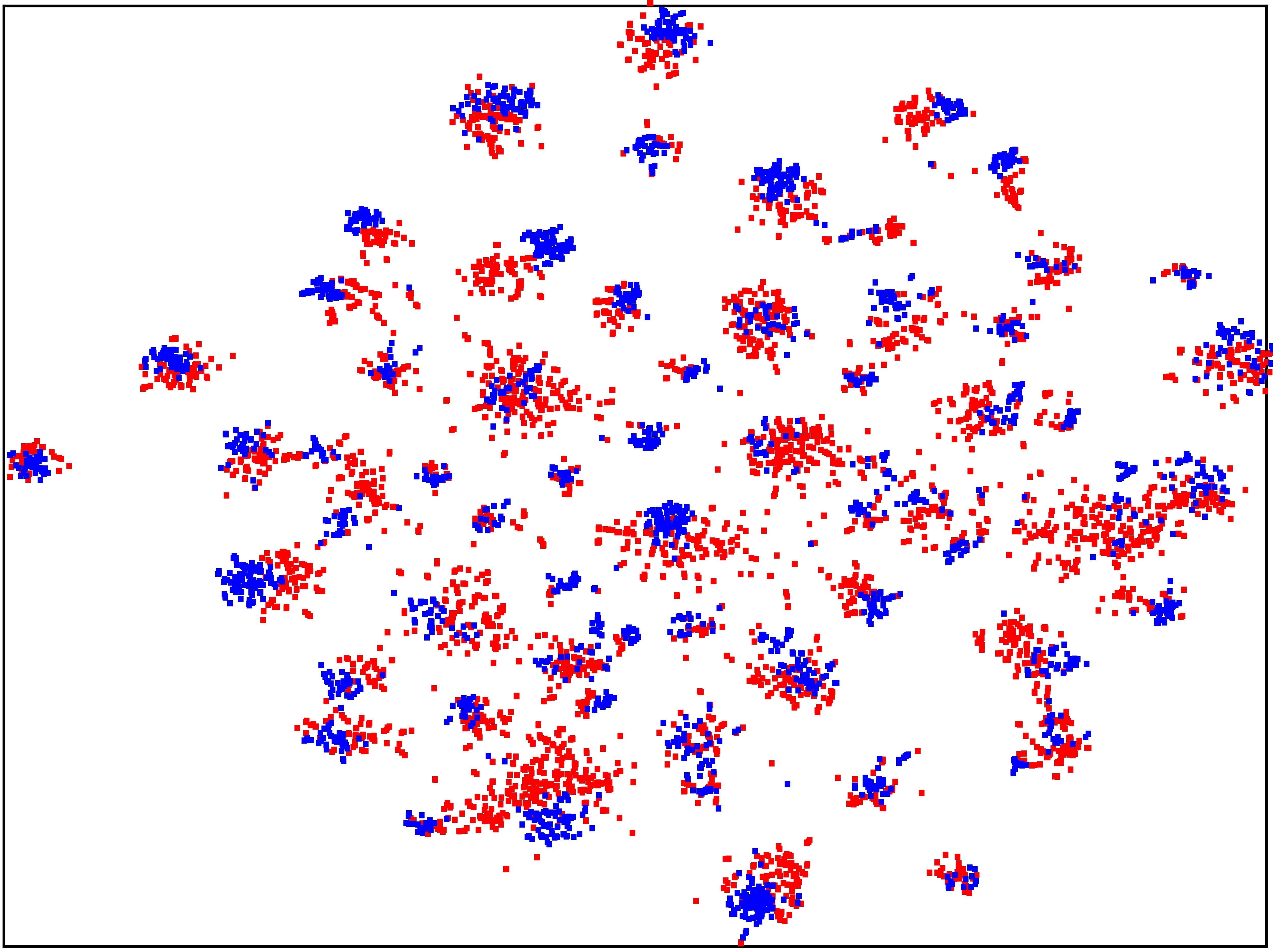}
					\label{tsne1}}
			\end{minipage}
			\begin{minipage}{.24\textwidth}
				\subfigure[BNM]{
					\includegraphics[width=0.9\textwidth]{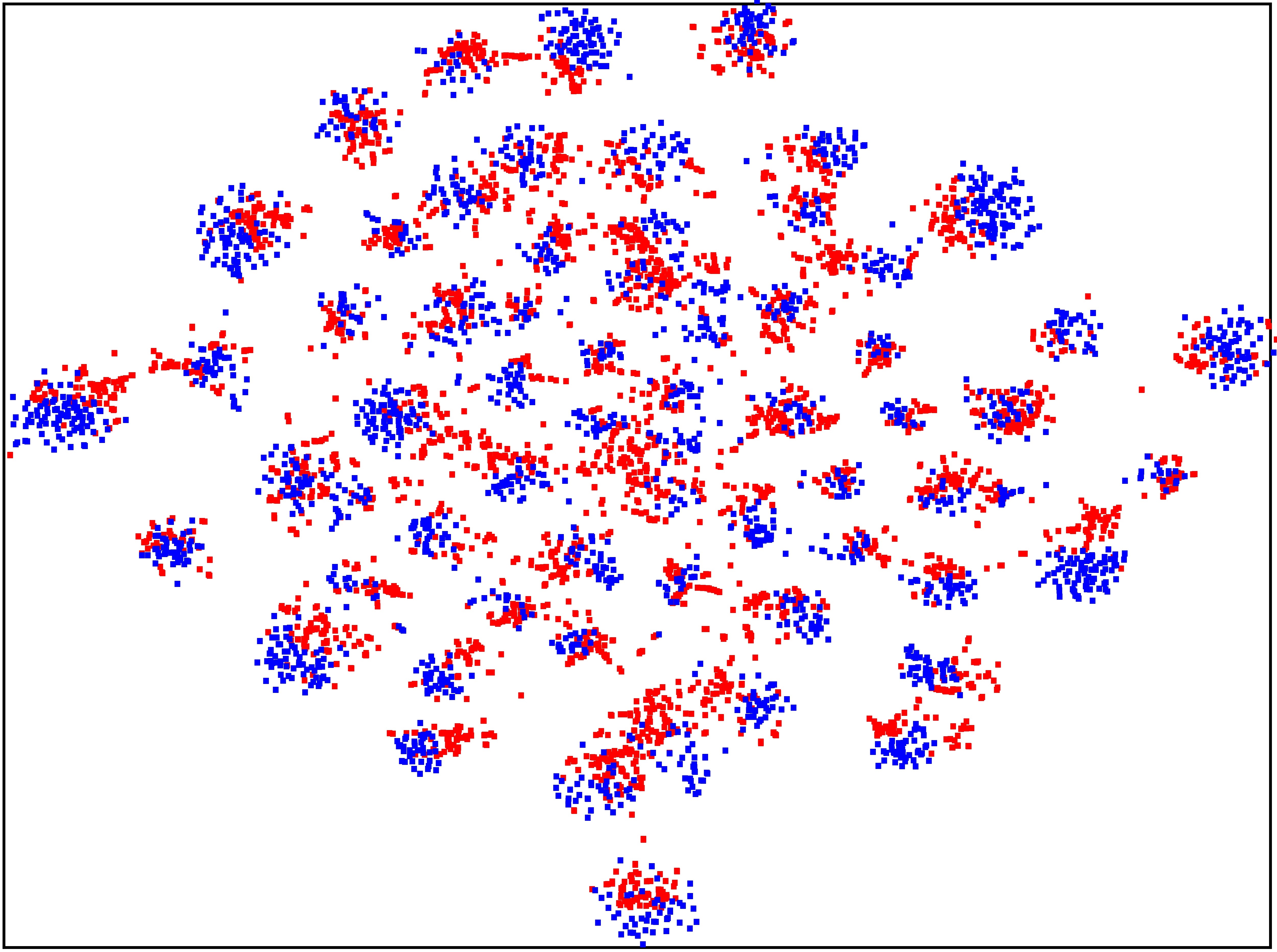}
					\label{tsne2}}
			\end{minipage}
			\begin{minipage}{.24\textwidth}
				\subfigure[BNM$^2$]{
					\includegraphics[width=0.9\textwidth]{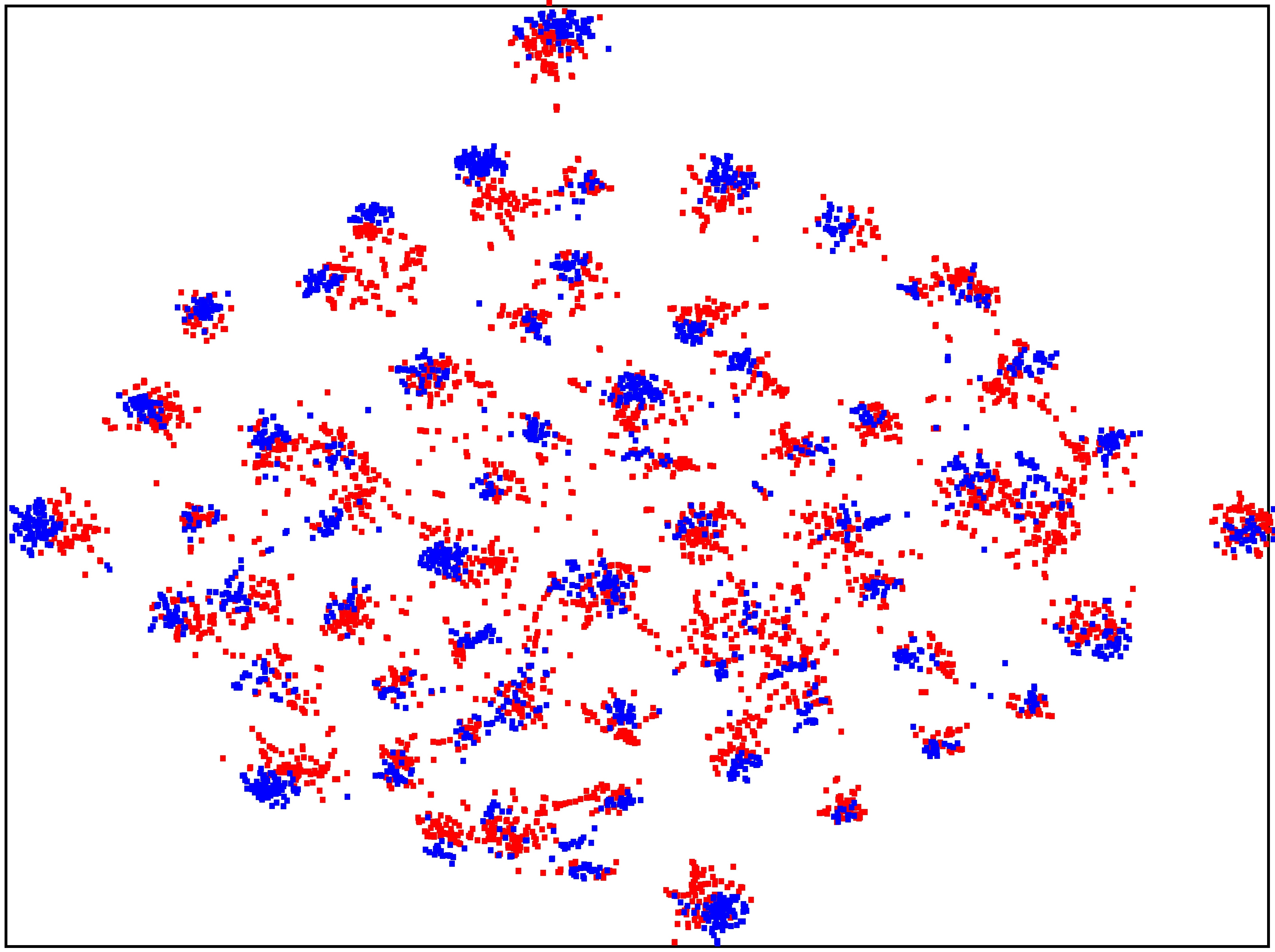}
					\label{tsne3}}
			\end{minipage}

			\vspace{-2ex}
			\caption{Visualization results of different methods with T-SNE on Ar $\rightarrow$ Cl. The source samples are in blue and target samples in red.}
			\label{tsne}
		\end{minipage}
	\end{center}
	\vspace{-3ex}
\end{figure*}


We also try adding BNM and BNM$^2$ to existing frameworks of CDAN~\cite{long2018conditional} and HDAN~\cite{cui2020hda}. CDAN+BNM outperforms CDAN and CDAN+EntMin by a large margin on both Office-31 and Office-Home. Besides, HDAN+BNM performs better than HDAN on Office-Home, which also shows that BNM cooperate well with other methods. BNM is more suitable to improve the performance of CDAN and HDAN on difficult tasks where the baseline accuracy is relatively low, such as D$\rightarrow$A and W$\rightarrow$A in Office-31, Ar$\rightarrow$Cl and Pr$\rightarrow$Cl in Office-Home. Collaborating with CDAN and HDAN, BNM$^2$ further outperform BNM on both Office-31 and Office-Home, which shows the necessity of BNMin. We also replace the methods of BNM and BNM$^2$ by FBNM and FBNM$^2$. The approximation methods FBNM and FBNM$^2$ achieve similar results compared with BNM and BNM$^2$, which again shows the effectiveness of the approximation. 

Recently, SHOT~\cite{liang2020we} achieves well-performed results, with similar constraint compared with BNM. We reproduce the results of BNM under the same environment with SHOT. Since source data is not utilized in the training of SHOT, we could not apply BNM$^2$ to the framework. We replace the constraint in SHOT on prediction diversity of information maximization by BNM, and FBNM, denoted as BNM-S and FBNM-S respectively. The reproduced results are shown in Table~\ref{tableoffice} and~\ref{tableofficehome}, which shows BNM-S could also outperform SHOT on average on both Office-31 and Office-Home. Specifically, BNM-S and FBNM-S could achieve obvious improvement than SHOT on difficult tasks, such as D$\rightarrow$A in Office-31 and Rw$\rightarrow$Cl in Office-Home. The results show that simply encouraging prediction discriminability and diversity by BNM is enough for most cases.

To further analyze the prediction discriminability of BNM, we calculate the typical measurement of entropy, as shown in Figure~\ref{entropy}. Compared with the baseline of source only, BNM achieves lower entropy but higher prediction discriminability on target domain, as shown in Figure~\ref{entropytarget} during most of the training process. On source domain, as shown in Figure~\ref{entropysource}, BNM$^2$ could reduce the prediction discriminability on source domain compared with BNM. Although the differences seem to be very tiny in the figure, the influence of BNMin is still significant, considering the explicit constraints of cross-entropy loss on source domain.

To validate that BNM could maintain diversity for domain adaptation, we show the diversity ratio in Office-Home on Ar $\rightarrow$ Cl and Ar $\rightarrow$ Pr in Figure~\ref{diversity}. Prediction diversity is measured by the average matrix rank, {\it i.e.}, average number of predicted categories in the randomly sampled batches. Thus the diversity ratio is measured by the predicted diversity dividing the average ground-truth number of categories. As shown in Figure~\ref{home1}, the diversity ratio of BNM is larger than that of EntMin by a large margin in Ar $\rightarrow$ Cl. The phenomenon is reasonable since EntMin encourage those samples near the decision boundary to be classified into the majority categories, leading to the reducing of the diversity in the batch. 
As shown in Figure~\ref{home2}, the diversity ratio of BNM is still larger than that of EntMin in Ar $\rightarrow$ Pr. But in Ar $\rightarrow$ Pr, the divergence between EntMin and BNM in diversity ratio is less than that of Ar $\rightarrow$ Cl. The smaller diversity differences correspond to fewer samples near the decision boundary due to the higher baseline accuracy in Ar $\rightarrow$ Pr. Thus BNM tends to be more effective on difficult tasks with rich data near the decision boundary. 

\begin{figure}[t]
	\begin{center}
		
		\begin{minipage}{.24\textwidth}
			\subfigure[Fixed Diversity]{
				\includegraphics[width=0.9\textwidth]{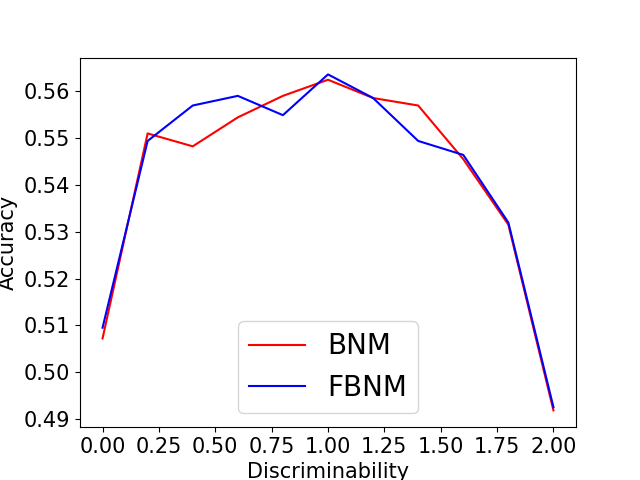}
				\label{retio1}}
		\end{minipage}
		\begin{minipage}{.24\textwidth}
			\subfigure[Fixed Discriminability]{	
				\includegraphics[width=0.9\textwidth]{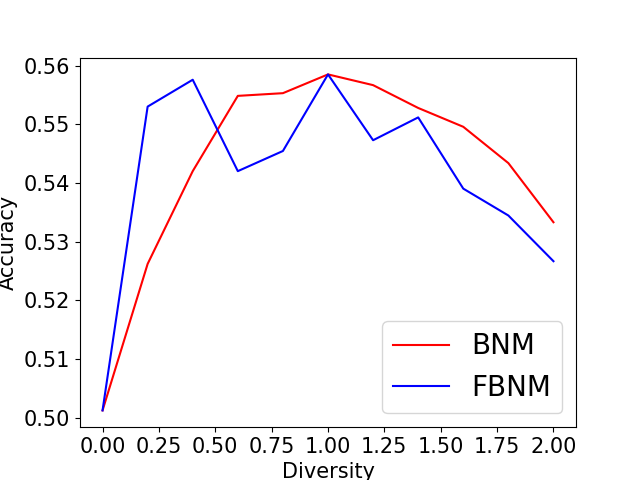}
				\label{retio2}}
		\end{minipage}
		\vspace{-3ex}
		\caption{Accuracy under different radio of discriminability and diversity. }
		\label{retio}
	\end{center}
	\vspace{-3ex}
\end{figure}

\begin{figure}[t]
	\begin{center}
		
		\begin{minipage}{.24\textwidth}
			\subfigure[Ar $\rightarrow$ Cl]{
				\includegraphics[width=0.9\textwidth]{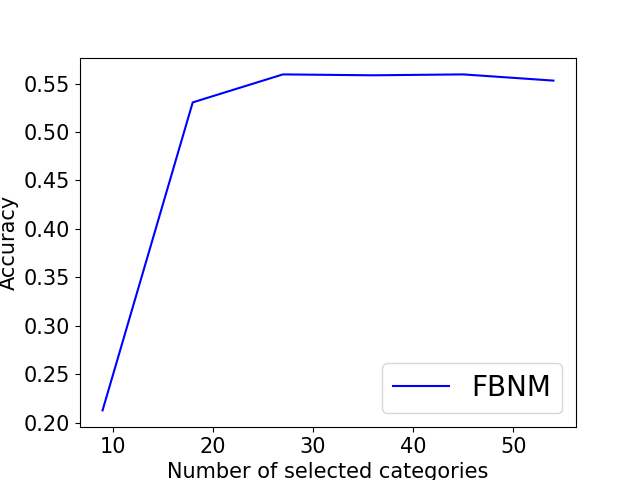}
				\label{numArCl}}
		\end{minipage}
		\begin{minipage}{.24\textwidth}
			\subfigure[Ar $\rightarrow$ Pr]{	
				\includegraphics[width=0.9\textwidth]{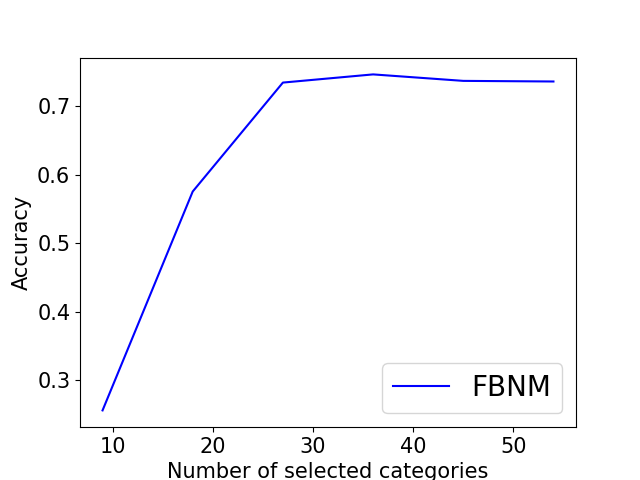}
				\label{numArPl}}
		\end{minipage}
		\vspace{-3ex}
		\caption{Accuracy under different selected number of categories with FBNM.}
		\label{numArClPl}
	\end{center}
	\vspace{-3ex}
\end{figure}

To analyze trade-off between improvement on prediction discriminability and diversity, we show the results with different ratios in Figure~\ref{retio}, by combining the objective functions of $F$-norm and nuclear-norm. $F$-norm denotes only discriminability while nuclear-norm denotes both discriminability and diversity. The weight of nuclear-norm is set to be $1$, meaning that the weight of diversity is fixed. The weight of $F$-norm varies from $-1$ to 1, so the range of discriminability is $[0,2]$. Under this setting (see Figure~\ref{retio}(a)), both BNM and FBNM could reach the optimal results when discriminability is $1$, corresponding to case that the weight of $F$-norm is $0$. 
In Figure~\ref{retio}(b), the sum of weights for nuclear-norm and $F$-norm is fixed to be $1$, meaning that the weight of discriminability is fixed. The ranges of the weights of nuclear-norm and $F$-norm are $[0,2]$ and $[-1,1]$, respectively.
Both BNM and FBNM reaches the optimal solutions when diversity is set to be $1$, also corresponding to the case that the weight of $F$-norm is $0$. The results show that only using the constraint of batch nuclear-norm is enough to achieve the optimal results, which corresponds to the equal weights of discriminability and diversity. The discriminability should not be encouraged too heavily by imposing additional discriminability constraint by $F$-norm.

We analyze the proper number of the selected categories for FBNM, {\it i.e.}, by varying $D$ in Eqn.~\ref{aproximate}. The accuracy under different $D$ for FBNM is shown in Figure~\ref{numArClPl}. 
In Sect.~\ref{fbnm-part}, following the formal definition $D=\min(B,C)$, the batch size $B$ is 36 and $C$ is 65, so $D$ is 36 to approximate the nuclear-norm. 
In experiments, this tends to be the best setting, since larger or small $D$ will damage the overall accuracy for both Ar $\rightarrow$ Cl and Ar $\rightarrow$ Pr. For example, in Figure~\ref{numArClPl}.(a), $D=18$ corresponds to 53.1 accuracy, with $2.7$ degradation compared with $D=36$. The results will approximate the optimal if $D$ is larger than 30. 

We also visualize the T-SNE results of Ar $\rightarrow$ Cl in Figure~\ref{tsne}, with source domain in blue and target domain in red. The visualization shows the large gap between source and target domain for Baseline trained only on source domain. The large domain gap is reduced by increasing prediction discriminability with BFM. For BNM, the visualization shows more red points near the blue points, with higher prediction diversity. BNM outperforms BFM on the overall accuracy in most tasks, which shows the importance of diversity. However, BNM still suffers from the problem that the samples from source and target domains are not tightly connected. BNM$^2$ mitigates this problem by reducing the discriminability on source domain, so the average distance between blue and red points is smaller.


\subsection{Balanced Domainnet}
To show the necessity of enforcing diversity and discriminability even on balanced dataset, we construct Balanced Domainnet where each category contains identical number of samples selected from Domainnet~\cite{peng2019moment}. In Balanced Domainnet, there are four domains including Real (R), Clipart (C), Painting (P) and Sketch (S) with large domain discrepancy. Each domain holds total 126 categories, with at least 50 training samples for each category. To maintain the balance for all the categories, different categories contain the same number of samples for one domain. The transfer tasks are selected by 7 domain adaptation scenarios, following \cite{saito2019semi}.
We list and compare the numbers of category samples in existing datasets, see Table 2 in Sect. 2.3 in Supplementary. We could find that existing datasets of Office-31, Office-Home, Domainnet and Semi Domainnet contain imbalanced categories, where Std monotonically increase with Mean. Typically, in Clipart (train) of Domainnet, the largest/smallest number of samples in a category is 328/8. This drastic dataset imbalance would inevitably result in prediction bias and model degradation.



\begin{table}[t]
	\begin{center}
		\vspace{-5pt}
		
		\caption{Calculation time in seconds~(1000 times) under different $B$ and $C$.}
		\vspace{-8pt}
		\label{timetable}
		\setlength{\tabcolsep}{1.3mm}{
			\begin{tabular}{ccccccc}
				\hline
				\hline
				$B$  & $100$ & $100$ & $ 100$ & $1000$ & $10000$  & $1000$  \\
				$C$  & $100$ & $1000$ & $10000$ & $100$ & $100$  & $1000$  \\
				\hline
				\hline
				EntMin~\cite{grandvalet2005semi}	&\textbf{0.05} 	&0.12 &1.80	&0.12 &1.64	&1.69 	 	 \\
				\hline
				\hline
				BNM	&0.89 	&2.72 	 &18.62	&1.93&6.39 &105.54	 	 \\
				FBNM &\textbf{0.05} 	&\textbf{0.11} 	&\textbf{0.82} &\textbf{0.08} &\textbf{0.30}  &\textbf{1.49}	 	\\
				
				\hline
				BNM / FBNM &16.34 	&23.88 &22.66 	&24.20 		&21.62 &70.78 	 \\ 
				\hline
				\hline
		\end{tabular}}
		\vspace{-10pt}
	\end{center}
\end{table}

\begin{table*}[t]
	\begin{center}
		\caption{ Accuracies (\%) on Semi DomainNet subset of ResNet34-based SSDA methods. Results with $*$ indicate the model training is not converged.}
		
		\label{SSDA}
		\vspace{-5mm}
		\resizebox{\textwidth}{!}{%
			\begin{threeparttable}
				\centering
				\setlength{\tabcolsep}{0.3mm}{
					\begin{tabular}{ccccccccccccccccc}
						\hline 
						
						\multirow{2}{*}{Methods} & \multicolumn{2}{c}{R$\rightarrow$C} & \multicolumn{2}{c}{R$\rightarrow$P} & \multicolumn{2}{c}{P$\rightarrow$C} & \multicolumn{2}{c}{C$\rightarrow$S} & \multicolumn{2}{c}{S$\rightarrow$P} & \multicolumn{2}{c}{R$\rightarrow$S} & \multicolumn{2}{c}{P$\rightarrow$R} & \multicolumn{2}{c}{Avg}\\
						
						&{1$_{shot}$} &{3$_{shot}$} &{1$_{shot}$} &{3$_{shot}$} &{1$_{shot}$} &{3$_{shot}$} &{1$_{shot}$} &{3$_{shot}$} &{1$_{shot}$} &{3$_{shot}$} &{1$_{shot}$} &{\small{3-shot}} &{1$_{shot}$} &{3$_{shot}$} &{1$_{shot}$} &\multicolumn{1}{c}{3$_{shot}$} \\
						
						\hline \hline

						\multicolumn{1}{c}{ResNet34 \cite{he2016deep}} &55.6&60.0  &60.6&62.2   &56.8&59.4  &50.8&55.0  &56.0&59.5  &46.3&50.1 &71.8&73.9 &56.9&\multicolumn{1}{c}{60.0}  \\
						
						\multicolumn{1}{c}{DANN~\cite{ganin2016domain}} &58.2&59.8  &61.4&62.8   &56.3&59.6  &52.8&55.4  &57.4&59.9  &52.2&54.9 &70.3&72.2 &58.4&\multicolumn{1}{c}{60.7}  \\
						
						\multicolumn{1}{c}{ADR~\cite{saito2017adversarial}} &57.1&60.7  &61.3&61.9   &57.0&60.7  &51.0&54.4  &56.0&59.9  &49.0&51.1 &72.0&74.2 &57.6&\multicolumn{1}{c}{60.4} \\
						
						\multicolumn{1}{c}{CDAN~\cite{long2018conditional}} &65.0&69.0  &64.9&67.3   &63.7&68.4  &53.1&57.8  &63.4&65.3  &54.5&59.0 &73.2&78.5 &62.5&\multicolumn{1}{c}{66.5} \\
						
						\multicolumn{1}{c}{EntMin~\cite{grandvalet2005semi}} &65.2&71.0  &65.9&69.2   &65.4&71.1  &54.6&60.0  &59.7&62.1  &52.1&61.1 &75.0&78.6 &62.6&\multicolumn{1}{c}{67.6} \\
						
						\multicolumn{1}{c}{MME~\cite{saito2019semi}} &70.0&72.2  &67.7&69.7   &69.0&71.7  &56.3&61.8  &64.8&66.8  &61.0&61.9 &76.1&78.5 &66.4&\multicolumn{1}{c}{68.9} \\
						\multicolumn{1}{c}{GVBG~\cite{cui2020gvb}}&70.8	&	73.3&65.9&	68.7	&71.1 &	72.9	&62.4&	65.3 	&65.1 &		66.6	&67.1 &	68.5	&76.8	&79.2	&68.4	&	70.6\\
						
						\hline
						\multicolumn{1}{c}{BNM} &70.6 	&72.7 	&69.1 	&70.2 	&71.0 	&72.5 	&61.3 	&63.9 	&67.2 	&68.8 	&61.0 	&63.0 	&78.2 	&80.3 	&68.3 	&70.2 \\
						\multicolumn{1}{c}{BNM$^2$} &71.4 	&72.8 	&69.3 	&70.5 	&70.8 	&73.0 	&61.3 	&64.1 	&67.4 	&68.7 	&61.4 	&63.4 	&78.9 	&80.5 	&68.6 	&70.4 \\
						\multicolumn{1}{c}{FBNM} &71.1 	&72.6 	&69.2 	&70.3 	&70.8 	&72.9 	&61.3 	&64.3 	&67.2 	&68.9 	&61.3 	&63.4 	&78.9 	&80.4 	&68.5 	&70.4 \\
						\multicolumn{1}{c}{FBNM$^2$} &71.0 &73.2 	&69.2 	&70.6 	&71.1 	&72.9 	&61.0 	&64.8 	&67.3 	&68.9 	&60.9 	&63.3 	&78.9 	&80.4 	&68.5 	&70.6 \\ 
						\hline
						\multicolumn{1}{c}{HDAN~\cite{cui2020hda}} &71.7 	&74.1 	&70.3 	&71.6 	&71.0 	&73.6 	&62.4 	&64.5 	&68.2 	&69.5 	&63.3 	&65.5 	&79.1 	&80.9 	&69.4 	&71.4 \\
						\multicolumn{1}{c}{HDAN+BNM} &\textbf{72.5} 	&\textbf{74.5} 	&70.8 	&71.9 	&71.5 	&\textbf{74.3}	&\textbf{63.5} 	&64.7 	&68.8 	&70.1 	&63.8 	&\textbf{66.1} 	&\textbf{80.3} 	&\textbf{81.4} 	&\textbf{70.2}	&71.8 \\
						\multicolumn{1}{c}{HDAN+BNM$^2$} &* 	&74.5 	&\textbf{70.9} 	&\textbf{72.1} 	&71.7 	&\textbf{74.3}	&* 	&65.0 	&68.9 	&70.1 	&* 	&65.8 	&* 	&\textbf{81.4} 	&* 	&\textbf{71.9}\\ 	
						\multicolumn{1}{c}{HDAN+FBNM} &71.8 &74.0 	&\textbf{70.9} 	&72.0 	&\textbf{72.0} 	&73.6 	&63.2 	&\textbf{66.5}	&69.0 	&\textbf{70.5} 	&\textbf{64.2} 	&65.8 	&79.8 	&81.2 	&70.1 	&\textbf{71.9}\\
						\multicolumn{1}{c}{HDAN+FBNM$^2$} &72.0 &73.6 	&70.8 	&72.0 	&71.9 	&73.7 	&63.3 	&66.3 	&\textbf{69.1} 	&70.4 	&\textbf{64.2} 	&65.9 	&79.8 	&81.3 	&\textbf{70.2} 	&\textbf{71.9}\\
						
						\hline
				\end{tabular}}
				\renewcommand{\labelitemi}{}
				\vspace{-5pt}
			\end{threeparttable}
		}
	\end{center}
	\vspace{-10pt}
\end{table*}


We show the results on Balanced Domainnet in Table~\ref{balancedomainnet}. Baseline1 refers to model trained with only source data, and Baseline2 refers to model trained with only source data along with the batch normalization (BN) layer updating using target unlabeled data. Updating BN layer could reduce the performance on source domain, and improve the performance on target domain. Directly increasing discriminability on target domain by BFM could achieve well-performed results. However, BNM could still outperform BFM under balanced situations. The results validates that promoting the prediction diversity is necessary even for balanced dataset. Meanwhile, compared with Baseline2, BNM could achieve higher performance on source domain, which also validates the effectiveness of BNM.

Considering the large gap between batch size $B=36$ and number of categories $C=126$ in Balanced Domainnet, we try different $K$ for BNM, {\it i.e.}, the number of batches used for calculating the nuclear-norm in Algorithm 1. 
Specifically, BNM perform well when $K$ is $2$ or $3$. Thus BNM could be fully explored when $K*B$ is larger or a bit smaller than $C$. Similar phenomenon by varying $K$ also appears on BNM$^2$, FBNM and FBNM$^2$, thus we set $K=3$ for all these variants.
BNM could also gain improved performance by combining with HDAN. However, BNM$^2$ does not outperform BNM on this balanced dataset. 
The results show that BNMin may be more suitable for imbalanced small-scale source domain, because models trained on balanced large-scale source domain are less likely to overfit. 
Finally, BNM performs well on difficult tasks, such as S$\rightarrow$P.

\begin{figure}[t]
	\begin{center}
		
		\begin{minipage}{.24\textwidth}
			\subfigure[Discriminability]{
				\includegraphics[width=0.9\textwidth]{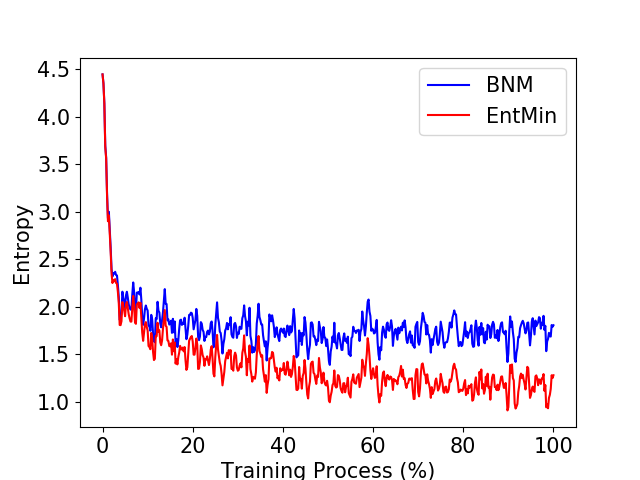}
				\label{balancediscriminability}}
		\end{minipage}
		\begin{minipage}{.24\textwidth}
			\subfigure[Diversity]{	
				\includegraphics[width=0.9\textwidth]{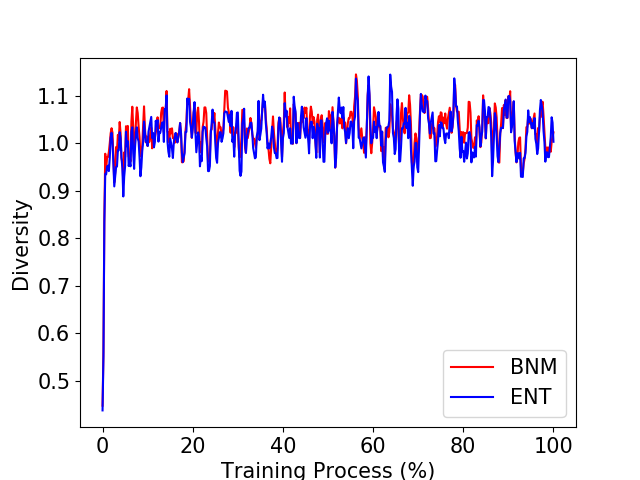}
				\label{balancediversity}}
		\end{minipage}
		\vspace{-3ex}
		\caption{Prediction discriminability and diversity for R$\rightarrow$C. Prediction discriminability is shown by entropy, while diversity is measured by average matrix rank dividing the average ground-truth number of categories. }
		\label{balanceentropy}
	\end{center}
	\vspace{-3ex}
\end{figure}

Under balanced situation, the prediction discriminability and diversity are shown in Figure~\ref{balanceentropy}. Both BNM and EntMin could increase discriminability as shown in Figure~\ref{balancediscriminability}. Since EntMin focuses on only discriminability, more discriminability could be achieved by EntMin. In contrast, BNM could still increase the prediction diversity compared with EntMin as shown in Figure~\ref{balancediversity}. In Table~\ref{balancedomainnet}, BNM achieves more accurate predictions than EntMin. The comparison shows that both prediction discriminability and diversity should be encouraged, even in balanced situation.

To show that how the approximation speeds up training, we calculate the training time in Table~\ref{timetable}. The results are based on Pytorch on desktop computer with CPU Intel(R) Core(TM) i7-4790K CPU @ 4.00GHz with 4 cpu cores. We run the results by randomly generating matrix $\mathrm{A}$, and the time for generating matrix $\mathrm{A}$ is removed from the calculation time. We calculate the total time on $1000$ random matrix $\mathrm{A}$. We denote the training time of BNM dividing FBNM as BNM/FBNM. The results show that FBNM is faster than EntMin and BNM for all situations. As an approximation method, FBNM could greatly reduce the calculation time of nuclear-norm, from $105.54$ used by BNM to $1.49$ for matrix size of $B=1000$ and $C=1000$. Also, we could identify the gradual improvement on BNM/FBNM, from $B=100$, $C=100$ to $B=1000$ and $C=1000$. The results show that FBNM could reduce the computation complexity by approximately $n$, which also validate that FBNM reduces the computation complexity from $O(n^3)$ to $O(n^2)$.


\subsection{Semi-supervised Domain Adaptation}
Semi-supervised domain adaptation (SSDA) is a more realistic setting, where domain adaptation is performed from labeled source data to partially labeled target data. In SSDA, we utilize Semi Domainnet, a typical dataset proposed in~\cite{saito2019semi}, which is selected from DomainNet~\cite{peng2019moment}. We take the same protocol with~\cite{saito2019semi} where the four domains including Real (R), Clipart (C), Painting (P) and Sketch (S) in total 126 categories. In this paper, following \cite{saito2019semi}, seven domain adaptation scenarios are formulated on the four domains, showing different levels of domain gap. All the methods are evaluated under the one-shot and three-shot settings as~\cite{saito2019semi}, where there are only $1$ or $3$ labeled samples per category in target domain.

We use ResNet34~\cite{he2016deep} as the backbones of the generator, optimized by SGD. The initial learning rate is $0.001$, with the momentum of $0.9$. BNM loss is combined with classification loss with $\lambda=0.5$. When BNM is regarded as a typical method, the source batch size is $24$ and target batch size is $48$, and $K=1$. When BNM is combined with existing methods, the source and target batch sizes are both $24$, and $K=3$. For fair comparison, we run three random experiments and report the average accuracy. 

The quantitative results on Semi Domainnet are summarized in Table~\ref{SSDA}. It is easily observed that BNM outperforms competitors under most settings, and achieves well performed results on average. The simple BNM method outperforms min-max entropy by a large margin, especially in difficult scenarios such as $1_{shot}$ and $3_{shot}$ C$\rightarrow$S. Meanwhile, to show the effect of BNM under SSDA, we also combine BNM with existing method of HDAN. The improvement on HDAN means that existing method might still lack enough prediction discriminability and diversity. 

Compared with the basic BNM, BNM$^2$ could further improve the performance on both $1$ shot and $3$ shot situations. Surprisingly, FBNM achieves better results compared with BNM, which shows the superiority of the approximated method. When cooperated with HDAN, BNM$^2$ could still slightly outperform BNM in most cases. However, in some scenarios such as $1_{shot}$ R$\rightarrow$C, the training process might not converge, denoted by $*$. The convergence issue results from the complex computation of singular value decomposition. Compared with original calculation of nuclear-norm in BNM, our approximated method could solve the problem of the convergence and performs efficiently.

\subsection{Unsupervised Open Domain Recognition}
\label{exp}

For unsupervised open domain recognition, we evaluate our methods on I2AwA~\cite{zhuo2019unsupervised}. In I2AwA, source domain consists of 2,970 images belonging to 40 known categories, via selecting images from ImageNet and Google image search engine. Target domain of I2AwA is AwA2~\cite{aw2} which contains 37,322 images in total. The target images belong to 50 categories, the same 40 known categories with source domain, and the remaining 10 individual classes as unknown categories. 

To obtain reliable initial task model on unknown categories, we construct the same knowledge graph for I2AwA with UODTN~\cite{zhuo2019unsupervised}. The graph structure is constructed according to Graph Convolutional Networks (GCN) \cite{kipf2016semi,zgcn}. The graph nodes include all categories of target domain along with the children and ancestors of the categories in WordNet~\cite{wordnet}. To obtain features of the nodes, we utilize the word vectors extracted by GloVe text model~\cite{GloVe} trained on Wikipedia. We use ResNet50~\cite{he2016deep} pretrained on ImageNet~\cite{deng2009imagenet} as our backbone. The parameters of the last fully connected layer are initialized by the parameters of GCN in the same categories.  

\begin{table}
	\caption{Accuracies (\%) and other measurements on I2AwA for ResNet50-based unsupervised open domain recognition methods.}
	\begin{center}
		\label{I2AwA}
		\vspace{-3pt}
		\setlength{\tabcolsep}{0.5mm}{
			\begin{tabular}{cccccccc}
				\hline               Method  &Discri.   &Diversity &Time  &Known        &Unknown         &All      &Avg     \\
				\hline\hline
				zGCN~\cite{zgcn}  &.&.   &.&77.2         &21.0            &65.0    &49.1    \\
				dGCN~\cite{adgpm}    &.&.  &.       &78.2        &11.6            &64    &44.9    \\
				adGCN~\cite{adgpm}    &.&.  &.      &77.3         &15.0           &64.1   &  46.2   \\
				bGCN~\cite{ubias}      &.&. &.      &84.6         &28.0           &72.6     &56.3  \\	
				pmdbGCN~\cite{pmd}     &.&.  &.    &84.7         &27.1           &72.5    &55.9  \\	
				UODTN~\cite{zhuo2019unsupervised}       &.&.  &.        &84.7 &31.7   &73.5&58.2\\
				
				Balance*~\cite{zhuo2019unsupervised}  &-0.759 &0.945 &1665.0		&85.9 &22.3	&72.4 &54.1\\	
				
				\hline
				EntMin &\textbf{-0.216} &0.839 &1786.3 &87.5&7.2	&70.5	&47.4\\
				BFM	&-0.260
				& 0.852 &1678.3&87.7 &9.2	&71.1	&48.4 \\				
				
				BNM        & -0.246 &0.951   & 1768.3 &\textbf{88.3} &{39.7}   &{78.0}&{64.0}\\
				BNM$^2$ &-0.398 &0.950 & 1779.4 &85.2 	&\textbf{49.5} &77.6	&{67.4} \\
				
				FBNM &-0.314 &\textbf{0.969}&\textbf{1661.9} &88.0 &43.6 &78.6  &65.8 \\
				FBNM$^2$ &-0.329 &0.962 &1663.0 &87.4 &49.2 &\textbf{79.3} &\textbf{68.3} \\

				\hline
			\end{tabular}
		}
	\end{center}
	\label{I2AwA}
	\vspace{-3ex}
\end{table}

\begin{figure*}
	\begin{center}
		\begin{minipage}{.98\textwidth}
			\begin{minipage}{.24\textwidth}
				\subfigure[All Category Accuracy]{	
					\includegraphics[width=0.9\textwidth]{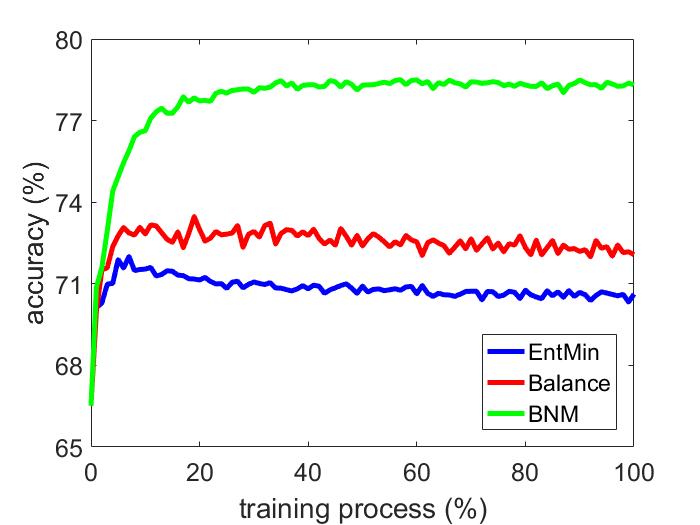}
					\label{compare1}}
			\end{minipage}
			\begin{minipage}{.24\textwidth}
				\subfigure[Known Category Accuracy]{
					\includegraphics[width=0.9\textwidth]{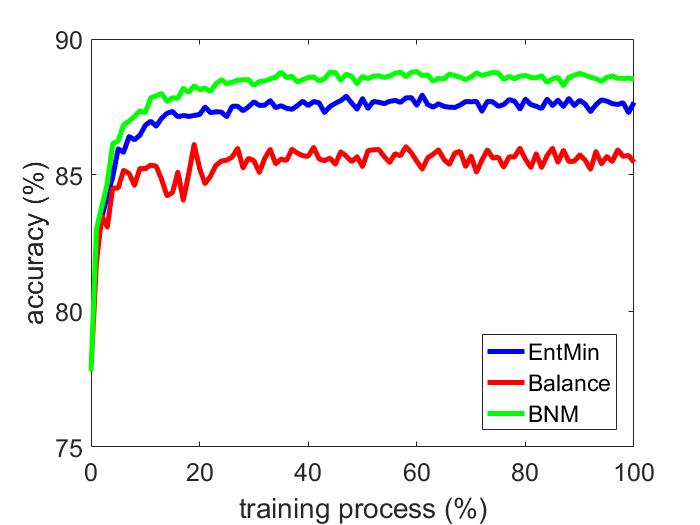}
					\label{compare0}}
			\end{minipage}
			\begin{minipage}{.24\textwidth}
				\subfigure[Unknown Category Accuracy]{
					\includegraphics[width=0.95\textwidth]{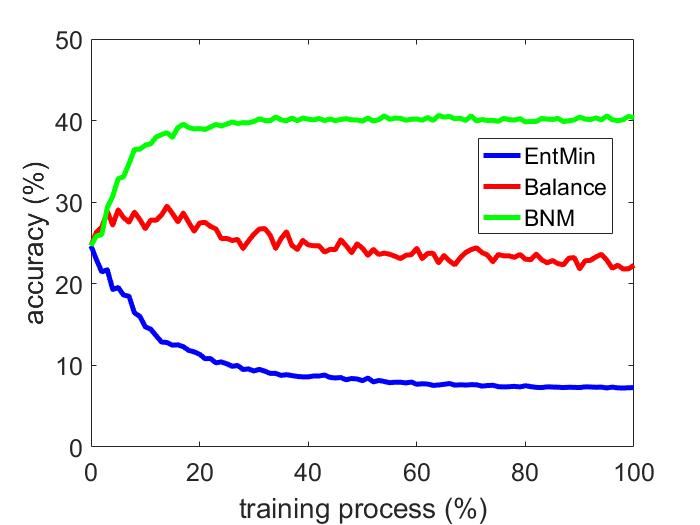}
					\label{compare3}}
			\end{minipage}
			\begin{minipage}{.24\textwidth}
				\subfigure[Unknown Category Ratio]{
					\includegraphics[width=0.9\textwidth]{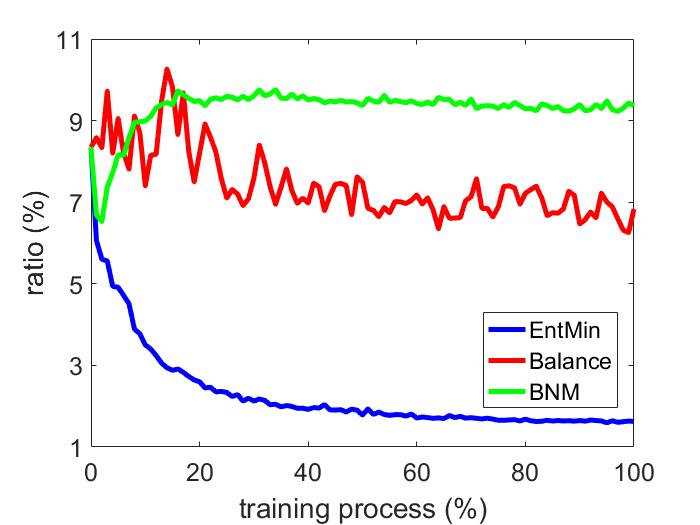}
					\label{compare2}}
			\end{minipage}

			
			\caption{Statistics for Entropy, Balance and Nuclear-norm in the whole training process for unsupervised open domain recognition.}
			\label{compare}
		\end{minipage}
	\end{center}
	\vspace{-2ex}
\end{figure*}

For fair comparison, we perform in the same environment with UODTN~\cite{zhuo2019unsupervised}. The experiments are implemented by Pytorch~\cite{torch}. We fix the batch size to be $48$ for both source and target domain. To train the network, we apply BNM on the classfication outputs on the total 50 categories for target domain, and minimize classification loss on the known 40 categories in source domain. Parameter $\lambda$ is set to be $2$ for BNM, FBNM and FBNM$^2$, while $1$ for BNM$^2$. We report the prediction results of known categories, unknown categories, all target categories, the average of known and unknown category accuracy, along with the measurements of prediction discriminability and diversity. The prediction discriminability (denoted as Discri. in short) is measured by negative entropy to ensure more discriminability with larger negative entropy. While prediction diversity is measured by the average predicted matrix rank dividing the average ground truth matrix rank. To show the advantage of approximation, we also calculate the training time for 6000 iterations on device of GeForce RTX 3090, denoted as Time. For each method, we run four random experiments and report the average result. 

\begin{table}
	\caption{Parameter Sensitivity on the I2AwA dataset for ResNet50-based unsupervised open domain recognition.}
	\begin{center}
		\setlength{\tabcolsep}{0.5mm}{
			\begin{tabular}{cccccccc}
				\hline                Method    &Discri.   &Diversity &Time  &Known        &Unknown         &All      &Avg    \\
				\hline	\hline
				BNM ($\lambda=1$) &0.340 &0.952 &1768.4	         &{88.0}&{39.4}   &{77.7}&{63.7}\\
				BNM ($\lambda=1.5$) &0.290 &0.956 &1770.2 
				&{88.1}&\textbf{39.7}   &{77.9}&{63.9}\\
				BNM ($\lambda=2$)  &0.246 &0.951 &1768.3 &\textbf{88.3}&\textbf{39.7}   &\textbf{78.0}&\textbf{64.0}\\
				BNM ($\lambda=3$) &0.208 &0.958 &1772.8 &{87.7} &{39.5}   &{77.5}&{63.6}\\
				BNM ($\lambda=4$) &\textbf{0.183} &\textbf{0.962} &\textbf{1767.4} &{87.4} &{38.6}   &{77.1}&{63.0}\\
				\hline
			\end{tabular}
		}
	\end{center}
	\label{stable}
	\vspace{-3ex}
\end{table}

The results are shown in Table~\ref{I2AwA}, we achieve remarkable improvement on I2AwA. In terms of prediction accuracy, we achieve 11.4\% improvement on the known categories compared with the baseline zGCN, and BNM surprisingly outperform zGCN by 19.0\% on the unknown categories. From the overall dataset, we achieve 13.3\% improvement on the whole dataset and 15.2\% improvement over zGCN. Besides, BNM outperforms the state-of-the-art UODTN~\cite{zhuo2019unsupervised} by 4.8\%. The results show that simple BNM is effective enough for unsupervised open domain recognition, which outperforms the combination of complex functions in UODTN. Besides, compared with BNM, BNM$^2$ could achieve higher average accuracy on known and unknown categories by producing more accurate predictions on unknown categories. Surprisingly, the approximated methods of FBNM and FBNM$^2$ could achieve even more accurate predictions on all accuracy and average category predictions, compared with BNM and BNM$^2$. Among our methods, FBNM$^2$ achieves the best results on both accuracy of All and Avg, with 10.1\% improvement on Avg and 5.8\% improvement on all compared with UODTN. 

Also, BNM could improve the measurements of discriminability and diversity. Compared with Balance, BNM could improve the prediction discriminability from $-0.759$ to $-0.246$. Though EntMin could ensure prediction discriminability, excessively encouraging discriminability seems to be unsatisfying. In terms of prediction diversity, the diversity ratio is improved from $0.839$ from EntMin to $0.969$ from BNM. We also find that the approximation method could achieve better prediction results by ensuring prediction diversity. From BNM to FBNM, diversity ratio changes from $0.951$ to $0.969$, with all accuracy from $78.0$ to $78.6$ and average accuracy from $64.0$ to $65.8$. In terms of the training time, the approximated methods demonstrate advantages of the training time, with reduction from $1768.3$ to $1661.9$ seconds by replacing BNM with FBNM.

We show the parameter sensitivity experiments in Table~\ref{stable}. The results show that BNM is relatively stable under different $\lambda$ on training time, known category accuracy and all accuracy. In terms of $\lambda$, the performance of all and avg reaches the highest when $\lambda=2$. Larger $\lambda$ means relatively higher prediction discriminability and diversity for target domain. Excessive prediction discriminability and diversity could reduce the knowledge learned from source domain, so $\lambda=2$ seems to be an appropriate setting.

We also compare the training process of EntMin, Balance constraint~\cite{zhuo2019unsupervised} and BNM in Figure~\ref{compare}. The prediction results on all categories, known categories and unknown categories are separately shown in Figure~\ref{compare1},~\ref{compare0} and~\ref{compare3}. BNM outperforms competitors on All accuracy, Known category accuracy and Unknown category accuracy in the whole training process. To explore the intrinsic effect of BNM on unknown categories, we show the unknown category ratio in Figure~\ref{compare2}, which is the ratio of samples in the target domain of I2AwA predicted into the unknown categories. Obviously, EntMin reduces the unknown category ratio by a large margin, which greatly damages the prediction diversity and accuracy on unknown categories. Though the unknown category ratio of BNM is reduced at first, it gradually raises along the training process, and after training it appears to be even higher than the initial ratio. This means BNM could increase prediction diversity by ensuring the ratio of predictions on minority categories. Though the Balance constraint could also keep the ratio of prediction on minority categories, the results of Balance loss tend to be unstable. Besides, the accuracy of Balance loss is much lower than BNM due to the lack of discriminability. The experimental phenomenon has steadily proved the effectiveness of BNM towards both discriminability and diversity.

\subsection{Discussion}

The chosen tasks in this paper are typical domain adaptation circumstances. Among the tasks and datasets, there are main differences in three aspects, {\it i.e.}, the domain discrepancy, category imbalance ratio and number of categories. First, there exists smaller domain discrepancy for Office-31, while large domain discrepancy for other datasets. Second, the categories are only balanced in Balanced Domainnet for UDA task. However, in Office-31 and Office-Home for UDA and semi-Domainnet for SSDA, the categories are highly imbalanced. Meanwhile, the I2AwA for UODR task is with extremely imbalanced category distributions, where some categories are even unseen in the source domain. Third, the number of categories is relative small for Office-31, Office-Home and I2AwA, but relatively large for Semi Domainnet and Balanced Domainnet.

From domain discrepancy aspect, BNM performs well in most cases, thus it is suitable for both large and small domain discrepancy, while experiments show that BNM could obviously improve the results when large domain discrepancy exists. 
From category imbalance aspect, BNM can perform well on strictly balanced dataset of Balanced Domainnet, and diverse types of imbalanced domain adaptation scenarios. Therefore, in both imbalanced and balanced situations, maintaining prediction diversity by BNM is necessary. Simple BNM even achieves state-of-the-art results for UODR task, thus BNM will perform better under extremely imbalanced situations. 

As the variants of BNM, BNM$^2$ outperforms BNM in most cases, except on Balanced Domainnet. The BNMin in BNM$^2$ tends to perform better under imbalanced small-scale circumstance. For BNM, the batch size should not be negligibly small compared to the number of categories. On Office-31, Office-Home and I2AwA, the number of categories is similar to the batch size, and BNM performs well under this situation. 
While on Balanced Domainnet and Semi Domainnet, it is required to enlarge the batch size by multiple batch optimization. Meanwhile, the fast method FBNM achieves similar results as BNM on all the tasks, and the model behaviors of FBNM (FBNM$^2$) and BNM (BNM$^2$) are quite similar on this two datasets.


\section{Conclusion}
    In this paper, we empharsize that prediction discriminability  and  diversity are the key to  determine the overall cross-domain generalization ability and robustness. We found theoretically that the prediction discriminability and diversity could be separately measured by the Frobenius-norm and rank of the batch output matrix. The nuclear-norm is the upperbound of the former, and the convex envelope of the latter.
    Accordingly, we propose Batch Nuclear-norm Maximization and Minimization, a  unified framework for visual domain adaptation, which performs batch nuclear-norm maximization on target domain and minimization on source domain. The two components play complementarily to adjust the discriminability and diversity of the visual models, and they can be combined with a wide range of existing domain adaptation frameworks.
    To compute BNM$^2$ more efficiently, we develop FBNM$^2$, accompanyied with multi-batch optimization, which achieves $O(n^2)$ computational complexity and more stable solution.
    Experiments validate the effectiveness and efficiency of our methods under three domain adaptation scenarios. 
    
    Future work are as follows. First, beyond image level classification tasks, we will adapt BNM$^2$ to other finer-grained visual understanding tasks such as semantic segmentation. 
    Second, we will explore how batch nuclear-norm will perform under other weakly supervised settings, {\it e.g.}, learning with noisy labels. 
    Finally, it would be interesting to investigate how BNM$^2$ will perform under more general out-of-distribution settings.



\textbf{Acknowledgement}. This work was supported in part by the National Key R\&D Program of China under Grant 2018AAA0102003, in part by National Natural Science Foundation of China: 62022083, 61672497, 61620106009, 61836002, 61931008 and U1636214, and in part by Key Research Program of Frontier Sciences, CAS: QYZDJ-SSW-SYS013.

\bibliographystyle{IEEEtran}
\bibliography{IEEEabrv,bare_jrnl}
%



\begin{IEEEbiography}[{\includegraphics[width=1in,height=1.25in,clip,keepaspectratio]{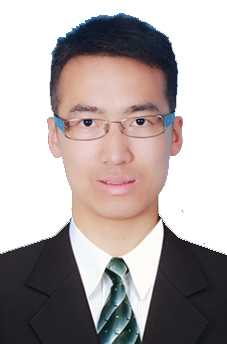}}]{Shuhao Cui} received the B.S. degree in department of automation from Tsinghua University, Beijing, China, in 2018. He is currently pursuing the M.S. degree in the Institute of Computing Technology, Chinese Academy of Sciences. His current research interests include machine learning, computer vision and transfer learning.
\end{IEEEbiography}

\begin{IEEEbiography}[{\includegraphics[width=1in,height=1.25in,clip,keepaspectratio]{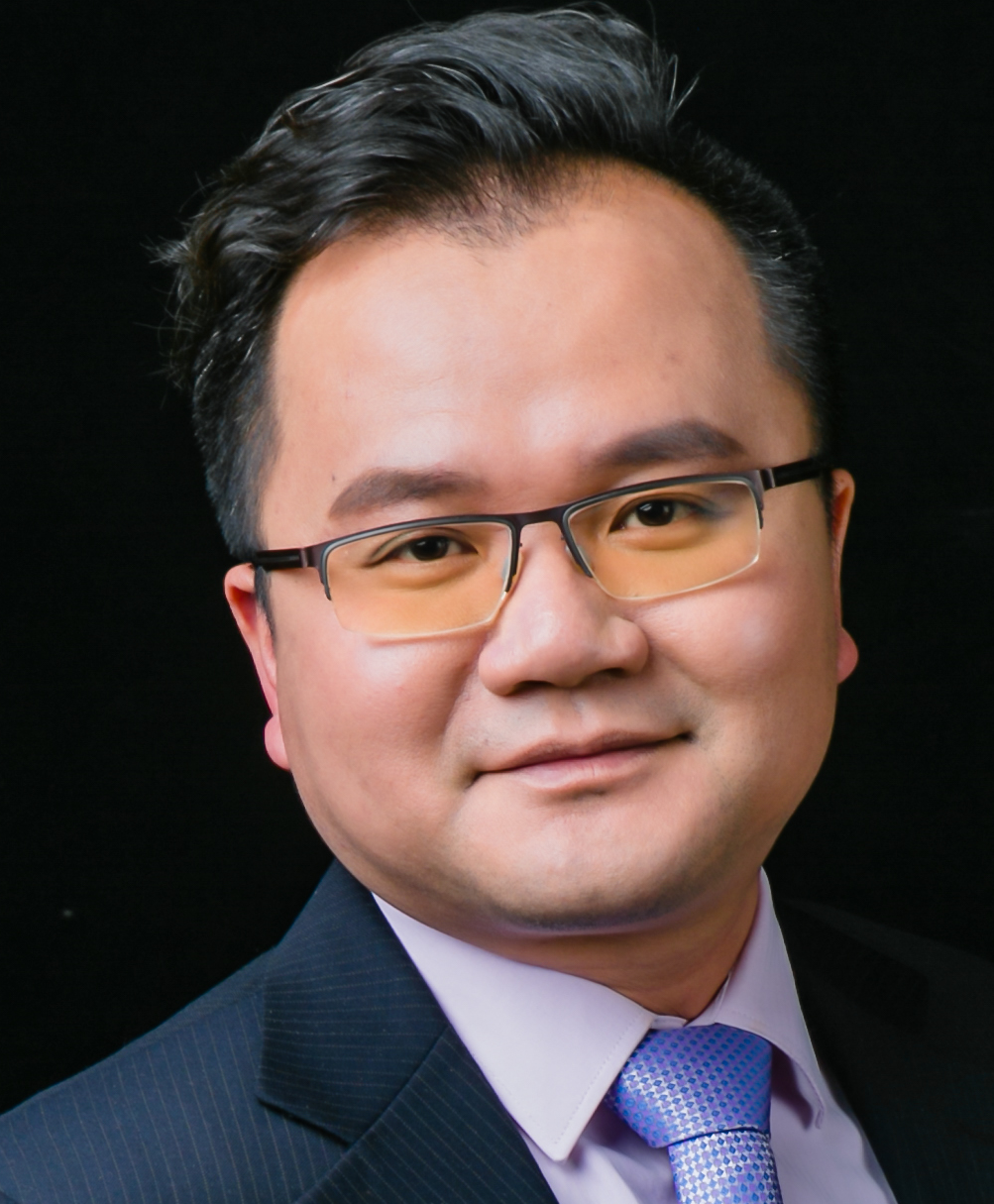}}]{Shuhui Wang} received the B.S. degree in electronics
	engineering from Tsinghua University, Beijing, China, in 2006, and the Ph.D. degree from the Institute of Computing Technology, Chinese Academy
	of Sciences, Beijing, China, in 2012. He is currently a Full Professor with the Institute of Computing Technology, Chinese Academy of Sciences.
	He is also with the Key Laboratory of Intelligent Information Processing, Chinese Academy of Sciences. His research interests include image/video understanding/retrieval, cross-media analysis and visual-textual knowledge extraction.
\end{IEEEbiography}

\begin{IEEEbiography}[{\includegraphics[width=1in,height=1.25in,clip,keepaspectratio]{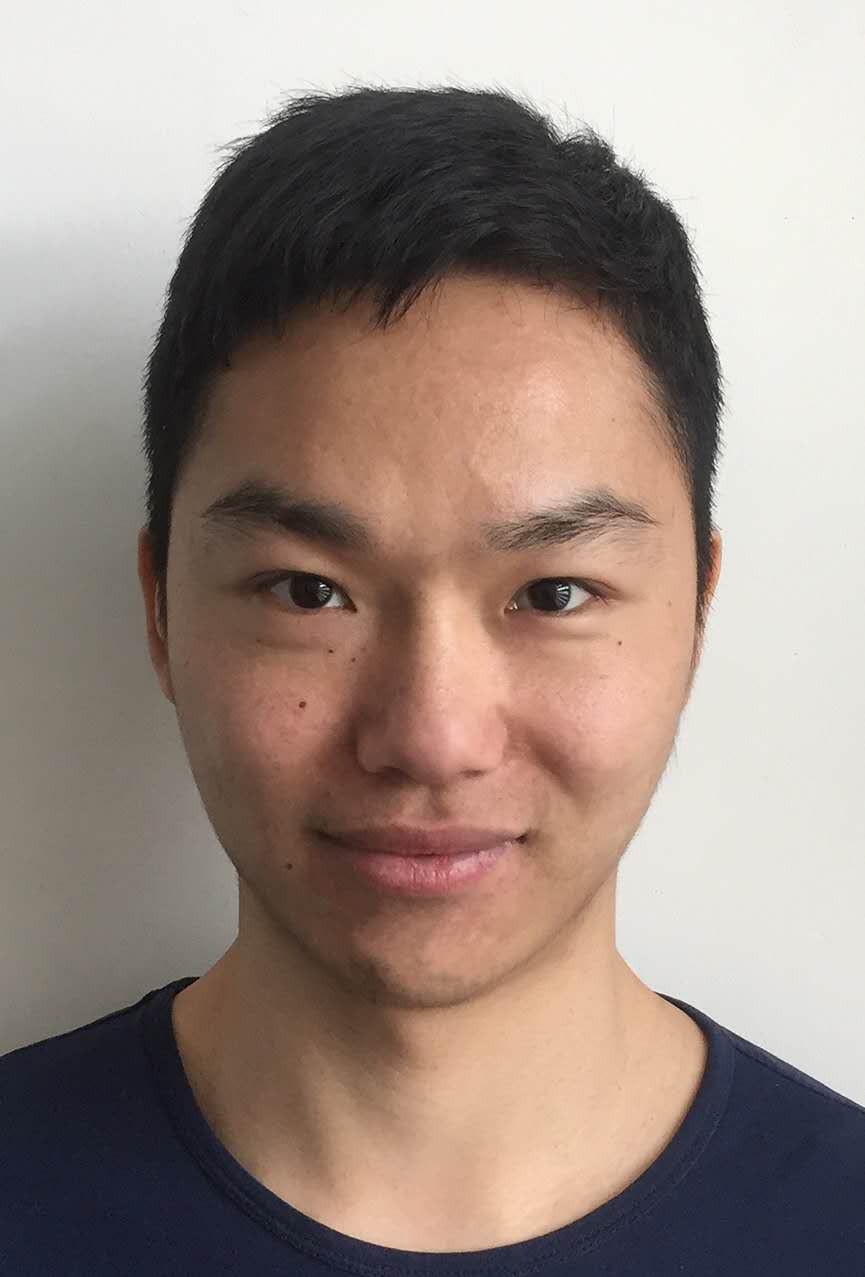}}]{Junbao Zhuo} received the B.S. degree in computer science from South China University of Technology, Guangzhou, China, in 2014 , and the Ph.D. degree from the Institute of Computing Technology, Chinese Academy  of Sciences, Beijing, China, in 2020.  He is currently a postdoctoral researcher with the Institute of Computing Technology, Chinese Academy of Sciences. His research interests include machine learning, deep learning and transfer learning.
\end{IEEEbiography}


\begin{IEEEbiography}[{\includegraphics[width=1in,height=1.25in,clip,keepaspectratio]{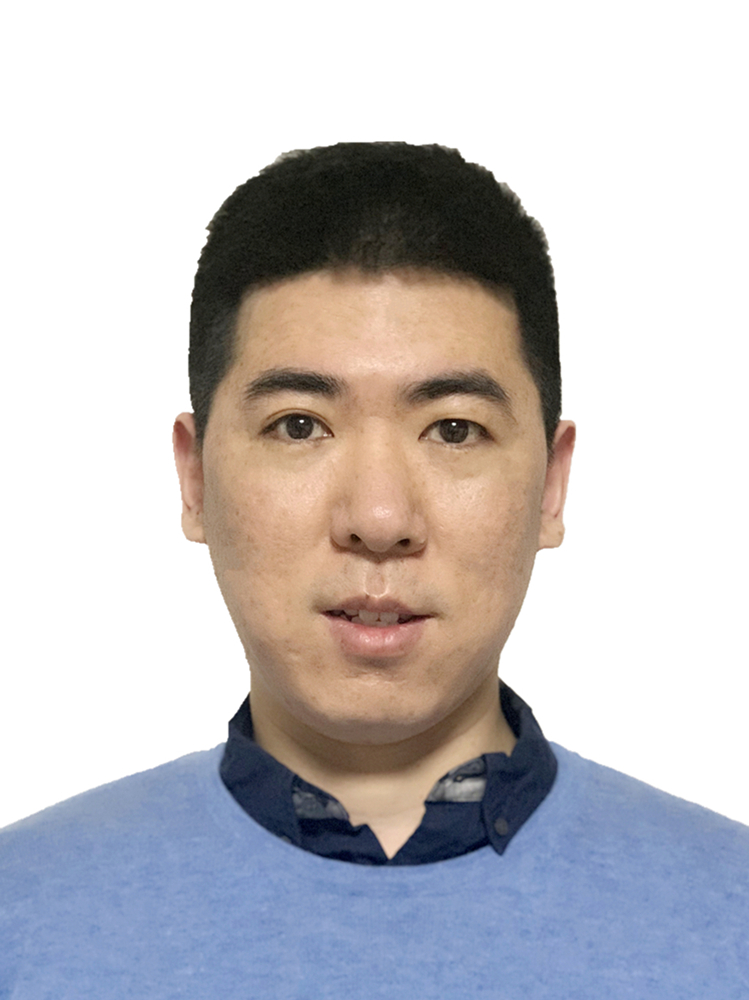}}]{Liang Li} received his B.S. degree from Xi’an Jiaotong University in 2008, and Ph.D degree from Institute	of Computing Technology, Chinese Academy	of Sciences, Beijing, China in 2013. From 2013 to 2015, he held a post-doc position with the Department of Computer and Control Engineering, University of Chinese Academy of Sciences, Beijing,	China. Currently he is serving as the associate	professor at Institute of Computing Technology, Chinese	Academy of Sciences. He has also served on a	number of committees of international journals and	conferences. Dr. Li has published over 60 refereed journal/conference papers. His research interests include multimedia content analysis, computer vision, and pattern recognition.
\end{IEEEbiography}

\begin{IEEEbiography}[{\includegraphics[width=1in,height=1.25in,clip,keepaspectratio]{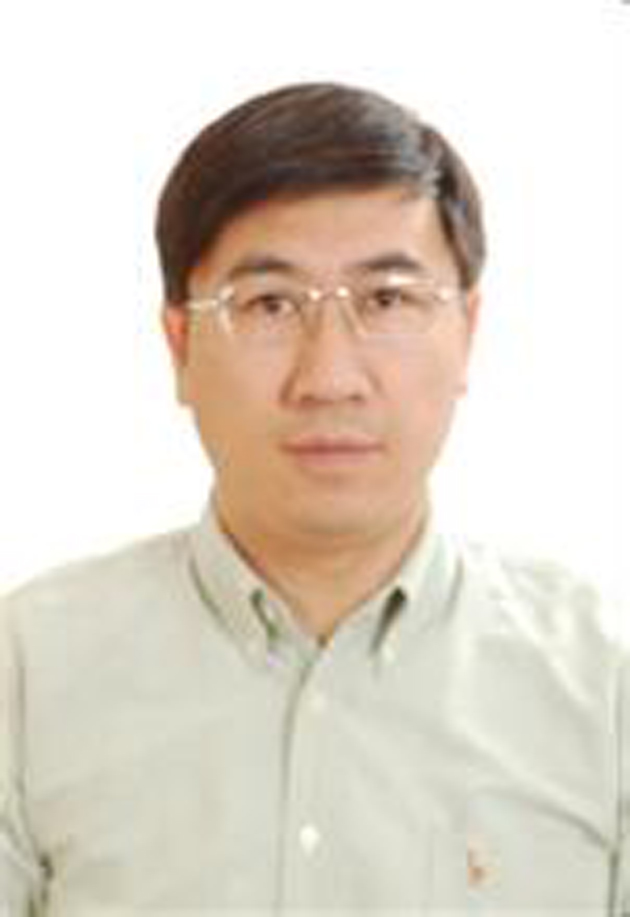}}]{Qingming Huang} received the B.S. degree
	in computer science and Ph.D. degree in computer engineering from the Harbin Institute of Technology, Harbin, China, in 1988 and 1994,
	respectively.
	He is currently a Chair Professor with the School of Computer Science and Technology, University of Chinese Academy of Sciences. He has authored over 400 academic papers in international journals, such as IEEE Transactions on Pattern Analysis and Machine Intelligence, IEEE Transactions on Image Processing, IEEE Transactions on Multimedia, IEEE Transactions on Circuits and Systems for Video Technology, and top level international conferences, including the ACM Multimedia,
	ICCV, CVPR, ECCV, VLDB, and IJCAI. He is the Associate Editor of IEEE Transactions on Circuits and Systems for Video Technology and the Associate Editor of Acta Automatica Sinica.
	His research interests include multimedia computing, image/video processing, pattern recognition, and computer vision.
\end{IEEEbiography}

\begin{IEEEbiography}[{\includegraphics[width=1in,height=1.25in,clip,keepaspectratio]{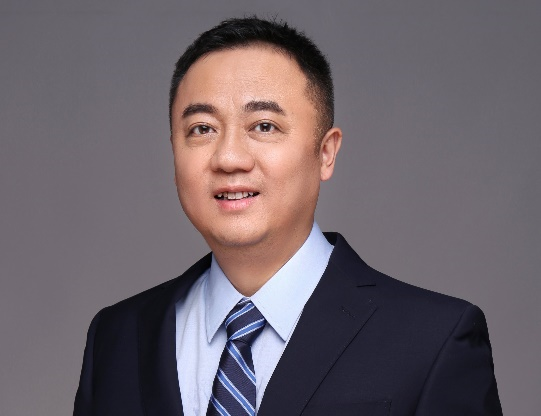}}]{Qi Tian} is currently a Chief Scientist in Artificial Intelligence at Cloud BU, Huawei. From 2018-2020, he was the Chief Scientist in Computer Vision at Huawei Noah's Ark Lab.  He was also a Full Professor in the Department of Computer Science, the University of Texas at San Antonio (UTSA) from 2002 to 2019. During 2008-2009, he took one-year Faculty Leave at Microsoft Research Asia (MSRA). 
	Dr. Tian received his Ph.D. in ECE from University of Illinois at Urbana-Champaign (UIUC) and received his B.E. in Electronic Engineering from Tsinghua University and M.S. in ECE from Drexel University, respectively. Dr. Tian's research interests include computer vision, multimedia information retrieval and machine learning and published 590+ refereed journal and conference papers. His Google citation is over 26100+ with H-index 78. He was the co-author of best papers including IEEE ICME 2019, ACM CIKM 2018, ACM ICMR 2015, PCM 2013, MMM 2013, ACM ICIMCS 2012, a Top 10\% Paper Award in MMSP 2011, a Student Contest Paper in ICASSP 2006, and co-author of a Best Paper/Student Paper Candidate in ACM Multimedia 2019, ICME 2015 and PCM 2007.
	Dr. Tian research projects are funded by ARO, NSF, DHS, Google, FXPAL, NEC, SALSI, CIAS, Akiira Media Systems, HP, Blippar and UTSA. He received 2017 UTSA President’s Distinguished Award for Research Achievement, 2016 UTSA Innovation Award, 2014 Research Achievement Awards from College of Science, UTSA, 2010 Google Faculty Award, and 2010 ACM Service Award. He is the associate editor of IEEE TMM, IEEE TCSVT, ACM TOMM, MMSJ, and in the Editorial Board of Journal of Multimedia (JMM) and Journal of MVA.  Dr. Tian is the Guest Editor of IEEE TMM, Journal of CVIU, etc. Dr. Tian is a Fellow of IEEE.
\end{IEEEbiography}

\end{document}